\documentclass[acmlarge]{acmart} 

\usepackage{array}
\usepackage{makecell}
\usepackage{xcolor}
\usepackage{pifont}
\definecolor{mygreen}{RGB}{56,142,60}
\newcommand{\cmark}{\textcolor{mygreen}{\ding{51}}}
\newcommand{\xmark}{\textcolor{red}{\ding{55}}}

\usepackage[final,commandnameprefix=ifneeded]{changes} 
\usepackage{subcaption}

\usepackage{graphicx}
\usepackage{subcaption}

\usepackage{adjustbox}

\usepackage[normalem]{ulem}

\usepackage{booktabs}
\usepackage{multirow}
\usepackage{graphicx} 

\AtBeginDocument{%
  }

\setcopyright{acmlicensed}
\copyrightyear{2026}
\acmYear{2026}
\acmDOI{3857931}

\acmJournal{IMWUT}
\acmVolume{10}
\acmNumber{4}
\acmArticle{213}
\acmMonth{12}

\title{\acronym: A Foundation Model for \added{Event-Based Behavioral Monitoring} in Smart-Homes}

\usepackage{pifont}

\author{Michele Fiori}
\correspondingauthor
\orcid{0009-0000-2462-3075}
\affiliation{
 \institution{University of Milan}
 \city{Milan}
 \country{Italy}}
\email{michele.fiori@unimi.it}

\author{Gabriele Civitarese}
\orcid{0000-0002-8247-2524}
\affiliation{
 \institution{University of Milan}
 \city{Milan}
 \country{Italy}}
 \email{gabriele.civitarese@unimi.it}

\author{Flora D. Salim}
\orcid{0000-0002-1237-1664}
\affiliation{
 \institution{University of New South Wales}
 \city{Sydney}
 \country{Australia}}
 \email{flora.salim@unsw.edu.au}

\author{Claudio Bettini}
\orcid{0000-0002-1727-7650}
\affiliation{
 \institution{University of Milan}
 \city{Milan}
 \country{Italy}}
 \email{claudio.bettini@unimi.it}

\newcommand{\acmlink}[2]{\href{#1}{\textcolor{ACMDarkBlue}{#2}}}
\authorsaddresses{Authors' Contact Information: \acmlink{https://orcid.org/0009-0000-2462-3075}{Michele Fiori} (corresponding author), University of Milan, Milan, Italy, \acmlink{mailto:michele.fiori@unimi.it}{michele.fiori@unimi.it}; \acmlink{https://orcid.org/0000-0002-8247-2524}{Gabriele Civitarese}, University of Milan, Milan, Italy, \acmlink{mailto:gabriele.civitarese@unimi.it}{gabriele.civitarese@unimi.it}; \acmlink{https://orcid.org/0000-0002-1237-1664}{Flora D. Salim}, University of New South Wales, Sydney, Australia, \acmlink{mailto:flora.salim@unsw.edu.au}{flora.salim@unsw.edu.au}; \acmlink{https://orcid.org/0000-0002-1727-7650}{Claudio Bettini}, University of Milan, Milan, Italy, \acmlink{mailto:claudio.bettini@unimi.it}{claudio.bettini@unimi.it}.}
 
\renewcommand{\shortauthors}{Fiori et al.}

\newcommand{\acronym}{DomusFM}

\begin{abstract}

\added{Smart-home sensor-based behavioral monitoring holds significant potential for healthcare, independent living, and early detection of functional or cognitive changes. In this setting, tasks like activity recognition, prediction, and pattern discovery provide complementary views of daily life, supporting the modeling of personal routines and habits, and their long-term changes. Existing approaches, however, face critical limitations. Supervised approaches require impractical amounts of labeled activity data to capture the variability of daily behavior across residents and environments. Foundation models represent a promising direction for learning transferable representations of latent behavioral patterns from sensor data. Still, current efforts are mostly designed for continuous inertial or physiological sensor data and do not address the sparse, discrete, and semantically rich event streams produced by smart homes.}
\added{In this paper, we introduce \acronym{}, a domain-specific foundation model for sensor-based behavioral monitoring in smart homes based on semantic event streams. \acronym{} employs a self-supervised dual contrastive learning paradigm to capture both event-level semantic attributes and sequence-level temporal dependencies. By integrating semantic embeddings from a lightweight language model and specialized encoders for temporal patterns and binary states, \acronym{} learns transferable representations that can be adapted across heterogeneous smart-home environments and tasks related to activity and event analysis.
Through a leave-one-dataset-out evaluation across seven public smart-home datasets, we demonstrate that \acronym{} consistently outperforms baselines on three downstream tasks: ADL recognition, next-k event prediction, and unsupervised clustering. \acronym{} has a small footprint and can be deployed on edge devices.}
\end{abstract}



\received{20 February 2007}
\received[revised]{12 March 2009}
\received[accepted]{5 June 2009}

\begin{document}
\setcopyright{cc}
\setcctype{by}
\acmJournal{IMWUT}
\acmYear{2026} \acmVolume{10} \acmNumber{4} \acmArticle{213}
\acmMonth{12} \acmDOI{10.1145/3857931}

\maketitle
\textcolor{red}{\textit{This paper has been accepted for publication in the Proceedings of the ACM on Interactive, Mobile, Wearable and Ubiquitous Technologies (IMWUT). For any inquiries, please contact michele.fiori@unimi.it}}
{\renewcommand{\thefootnote}{\ding{41}}\footnotetext{Corresponding author.}}

\section{Introduction}
Over the past decades, technological advancements have driven the widespread adoption of Internet of Things (IoT) devices that integrate a wide range of sensors capable of monitoring diverse phenomena, such as temperature, light intensity, pressure, energy consumption, and human movement \cite{babangida2022internet}. The availability of such rich sensor data has enabled the development of numerous context-aware smart-home applications, including healthcare monitoring and assistance \cite{bakar2015activity}. 


The growing demand for effective in-house \added{behavioral monitoring} and digital healthcare accelerated the diffusion of smart-home environments, driving the development of different analytical tasks \cite{alam2012review}. Among these, the most extensively studied is the recognition of Activities of Daily Living (ADL).
Other relevant tasks include next-events prediction for proactive healthcare assistance and domotics, \added{routine discovery}, and anomaly detection. \added{To date, most approaches rely on task-specialized deep learning techniques, which require large amounts of data (often labeled) from the target domain for effective training.} However, in the smart-home domain, data collection and annotation are particularly costly, time-consuming, and often impractical due to privacy concerns. 

More recently, transfer learning approaches have been proposed to leverage training from related domains or tasks \cite{cook2013transfer}. Foundation models represent the most recent paradigm of transfer learning, demonstrating the potential of models pretrained on broad data to adapt effectively to diverse downstream tasks with minimal supervision. In the context of behavioral modeling, two main paradigms have emerged: specialized foundation models trained on large-scale inertial sensor data from wearables \cite{zhang2024unimts, hong2024crosshar, yuan2024self, wei2025one}, and approaches that leverage existing large language models (LLMs) by encoding sensor events as text \cite{chen2024towards, civitarese2025large, fiori2025leveraging, cleland2024leveraging, cumin2025knowledge, cruciani2025few, fritsch2024hierarchical, thukral2025layout}. While the former have shown impressive results for continuous, high-frequency \added{sensor data (e.g., from inertial or physiological devices)}, they are not well-suited to the sparse, discrete, and semantically-rich nature of smart-home binary sensor events. \added{LLM-based approaches} face practical challenges, including the need for careful prompt engineering and reliance on either proprietary external APIs or costly hardware for local deployment. Importantly, these approaches repurpose foundation models trained on text corpora rather than developing foundation models specifically for smart-home sensor data. As a result, this domain still lacks a true foundation model pretrained at scale on diverse smart-home datasets.


\added{
In this paper, we introduce \acronym{}, a domain-specific foundation model for sensor-based behavioral monitoring in smart homes based on semantic event streams. \acronym{} operates on behaviorally meaningful events, including events produced by binary sensors and virtual events obtained by discretizing continuous sensor streams into semantic states. This event-level abstraction follows a common practice in smart-home behavior modeling~\cite{arrotta2022dexar}: raw sensor values are often device-specific and weakly aligned with the behavioral concepts of interest, while semantic events provide a common interface for heterogeneous smart-home sensing data while preserving the semantic structure relevant to activity and behavior analysis.}

\added{Unlike wearable-based sensor foundation models that target broader cross-modal or generative objectives by aligning sensor signals with language or other high-level modalities (e.g., IMU2CLIP, SensorLLM, SensorLM)~\cite{moon2023imu2clip,li2025sensorllm, zhang2026sensorlm}, \acronym{} adopts a domain-specific view centered on self-supervised pretraining, reusable representations, and transfer across heterogeneous sensing configurations over event streams captured by environmental sensors, similarly to recent foundation models for human sensing (e.g., MASTER, HAR-FM, SleepFM)~\cite{zhu2025master,bian2026foundation,qiu2025towards,thapa2026multimodal}. }

\acronym{} employs a two-stage self-supervised contrastive learning framework: the first stage learns event-level representations by modeling semantic sensor attributes, while the second stage contextualizes these representations by capturing temporal dependencies within event windows. The resulting pretrained backbone can then be adapted to downstream smart-home tasks without training the feature extractor from scratch.

\added{Compared to the existing solutions, \acronym{}'s architecture introduces the following novelties: i) encoding smart-home sensor events into semantic attributes (e.g., type, status, and temporal information) through specialized encoders, ii) modeling the relationship between these attributes thanks to attention mechanisms, and iii) for each event, generating a contextualized embedding also taking into account its previous and following events.
This design makes \acronym{} dataset-agnostic: events are represented through abstract semantic and temporal attributes rather than dataset-specific sensor identifiers or activity definitions. As a result, the model can be pretrained on heterogeneous smart-home datasets and learn behavioral patterns that transfer across sensor configurations, activity sets, and deployment settings.}

\added{We evaluate \acronym{} using a leave-one-dataset-out methodology across seven public datasets, where each target dataset is held out from pretraining and used only for downstream adaptation and testing. The evaluation covers three representative tasks commonly studied in smart-home behavioral monitoring~\cite{morita2023health, cao2010depth, cook2015activity}: ADL recognition, next-k event prediction, and unsupervised clustering. The results show that \acronym{} provides effective transferable representations for semantic smart-home event streams, improving over task-specific baselines across heterogeneous environments.}

To summarize, our contributions are the following:
\begin{itemize}
    \item We introduce \acronym{}, the first \added{lightweight} foundation model specifically designed and pretrained for \added{Event-Based Behavioral Monitoring in Smart-Homes.}
    
    
    \item \added{By encoding smart-home sensor events into semantic attributes, 
    \acronym{} adapts a dual contrastive pretraining strategy for dataset-agnostic representation learning. In particular, we introduce an attribute-level loss that captures relationships among each event's semantic attributes, and an event-level loss that models temporal dependencies across events.}
    \item Through extensive experiments across seven diverse state-of-the-art datasets and \added{three} downstream tasks, we show that \acronym{} substantially outperforms existing approaches, particularly under realistic data scarcity conditions. Our leave-one-dataset-out evaluation rigorously validates the model's ability to generalize to completely unseen smart-home environments.
\end{itemize}
As an additional contribution, we published \acronym{} as open-source software \footnote{\url{https://gitlab.di.unimi.it/ewlab/public/domusfm}}.


\section{Related Work}
\label{sec:related_work}

\subsection{Foundation Models}
\label{def:foundation_model}

\subsubsection{\added{Definition}}

\added{
To date, no universally accepted formal 
definition of a \emph{foundation model} exists in the literature. A common definition~\cite{bommasani2021opportunities,schneider2024foundation, caballar2024what, choi2025defining}, describes a foundation model as \textit{any model that is trained on broad data (generally using self-supervision) that can be adapted (e.g., fine-tuned) to a wide range of downstream tasks}. This work builds on this widely accepted conceptual definition and, in particular, formalizes it as follows:
A \textbf{foundation model} is a function $g_\theta$ with trainable parameters $\theta$ that satisfies the following properties:}
\begin{enumerate}
    \item \added{\textbf{Broad pretraining:} $g_\theta$ is trained on a large and diverse dataset $D$ using a general-purpose objective (e.g., self-supervised learning), capturing patterns applicable in a given domain.}
    \item \textbf{Adaptability:} $g_\theta$ can be adapted to a set of downstream tasks $\mathcal{T}_{d-stream}$, without re-training from scratch.
\end{enumerate}

\added{The same definition has been adopted in various subsequent works spanning a huge set of domains like, language~\cite{devlin2019bert, brown2020language, chowdhery2023palm}, vision-language~\cite{radford2021learning, yuan2021florence, wang2023image}, medical-images~\cite{liang2025vision, zhang2024challenges, amadou2024echoapex, kim2025echofm}, acoustic respiratory model~ \cite{zhang2024towards}, weather and climate~\cite{schmudefoundation}, or protein analysis \cite{verkuil2022language, bjerregaard2025foundation}.
Foundation models have recently been extended to time series, learning general-purpose representations transferable across tasks~\cite{jin2023time, rasul2023lag, garza2023timegpt, cao2023tempo, chang2025llm4ts, abbaspourazad2023large}.
}

\subsubsection{Foundation Models in Sensor-Based Behavioral Monitoring}
\label{subsec:fm_4_har}

\added{
The majority of existing foundation models in sensor-based behavioral monitoring target continuous, high-frequency, and often fixed-rate sensor streams, mainly leveraging IMU sensors~\cite{zhang2024unimts, hong2024crosshar, yuan2024self, wei2025one} and physiological sensors~\cite{thapa2026multimodal, gu2026cardiac,pillai2025papagei}. The considered downstream tasks include human activity recognition, sleep quality estimation or prediction, emotion recognition, and health-related inference tasks.
}

\added{
Some approaches combine time-series processing with LLMs to exploit their semantic knowledge, either by transforming time series into tokenized or textual representations, or by coupling temporal encoders with LLM backbones. This enables the use of semantic priors and contextual reasoning for sensor-based inference~\cite{zhang2026sensorlm,li2025sensorllm, yan2025large, zhang2025large}.
} 

\added{
These approaches have limited applicability in smart-home environments, where signals mostly come from environmental sensors and observations are typically represented as discrete sensor events separated by irregular time gaps. 
}

\added{
For this reason, the smart-home literature has mainly investigated LLMs as general-purpose foundation models for specific downstream tasks, such as ADL recognition. The rationale is that discrete sensor events are inherently associated with semantic information, such as object usage, room occupancy, or human-object interactions, which makes natural-language representations a plausible interface for LLMs. Accordingly, prompting-based approaches exploit zero-shot or in-context learning capabilities by encoding sensor data into textual descriptions~\cite{chen2024towards, civitarese2025large, fiori2025leveraging, xue2023promptcast, xue2023utilizing}. 
Fine-tuning strategies have also been explored to improve recognition performance~\cite{cleland2024leveraging, cumin2025knowledge, cruciani2025few, fritsch2024hierarchical}. 
}

\added{
Nevertheless, these methods require careful prompt design and nontrivial engineering to translate sensor events into natural language, and their performance can be highly sensitive to such design choices. Moreover, they typically rely on LLMs as task-oriented reasoning engines rather than learning sensor-native foundation representations for irregular, event-based smart-home data.
}

\added{
Unlike foundation models leveraging IMUs and physiological sensors and LLM-based methods, \acronym{} is a specialized foundation model for behavioral monitoring that adopts an event-based strategy, considering signals from both physical and virtual sensors in smart homes.
}


\subsection{Downstream Tasks in Smart-Home \added{Behavioral Monitoring}}
\label{subsec:downstream_tasks}

\added{
Smart-home behavioral monitoring aims to infer, model, and anticipate residents' routines from ambient sensor observations~\cite{cook2015activity}. Within this context, downstream tasks address complementary aspects of behavior understanding, including recognizing ongoing activities, forecasting future events, and discovering recurring behavioral patterns from unlabelled data. These tasks are central to the interpretation of daily routines and to the identification of regularities, deviations, or long-term behavioral changes that may be relevant for monitoring and care.}

\subsubsection{\added{Behavior recognition}}
\added{
Recognizing Activities of Daily Living (ADLs) is a core task in smart-home behavioral monitoring, as it maps sensor observations to semantic descriptions of residents' routines. Deep learning methods, including CNNs and LSTMs, have achieved strong performance in ADL recognition~\cite{liciotti2020sequential, wang2020activities}. However, they typically require large labelled datasets, which are difficult to collect in real homes due to annotation cost, privacy constraints, and the variability of residents' routines. Semi-supervised~\cite{abdallah2018activity}, transfer-learning~\cite{sanabria2021unsupervised, soleimani2021cross}, and self-supervised approaches~\cite{haresamudram2022assessing} have been proposed to mitigate this limitation, although most work has focused on wearable-based HAR rather than ambient smart-home environments. As a result, the effectiveness of self-supervised learning for smart-home ADL recognition remains substantially underexplored.}

\subsubsection{\added{Behavior forecasting}}
\added{
Behavior forecasting focuses on modeling the temporal dynamics of residents' routines in order to anticipate future sensor events, activities, or event sequences from past observations. Compared with behavior recognition, this task has received less attention in smart-home settings. Early probabilistic models~\cite{casagrande2019comparison} and RNN-based approaches~\cite{casagrande2018sensor} generally rely on supervised learning and often predict only the next event, which can be ambiguous and of limited practical utility when considered in isolation. More recent transformer-based models~\cite{takeda2024sensor} address this limitation by generating sequences of future events. However, effective forecasting under limited supervision and across heterogeneous home environments remains an open problem.}

\subsubsection{\added{Behavioral representation}}
\label{subsubsec:clustering_related}
\added{
Learning latent representations of human behavior is crucial for discovering recurring routines and behavioral patterns without requiring activity labels. This is particularly relevant in smart-home settings, where annotation is costly and residents' behavior may vary substantially across homes and over time. Such representations can also support healthcare-oriented tasks, including anomaly detection~\cite{fahad2021activity} and behavioral change detection~\cite{gupta2020tracking,civitarese2026x}, by enabling the identification of deviations or shifts in learned behavioral spaces. Existing representation-learning approaches in smart homes include autoencoders~\cite{chen2018latent}, NLP-inspired sensor embeddings~\cite{bouchabou2021using,hiremath2022bootstrapping}, and hybrid sequence- and image-based encodings~\cite{zhao2026care}. To the best of our knowledge, we are the first to propose a smart-home behavioral representation model that is pre-trained on multiple datasets to learn a generalized representation of smart-home events.}

\section{\added{A Conceptual Framework for Smart-Home Foundation Models}}
\added{In the following, we describe what we believe are fundamental components for an event-based smart-home foundation model.}

\subsection{\added{Sensor Event Stream and Segmentation}}
\label{subsec:sensor_event_stream_and_segmentation}
\added{
Let $S = \{s_0, s_1, \ldots, s_M\}$ be the set of $M$ sensors deployed in the environment. The smart-home environment generates a stream of discrete events $E = \{e_1, e_2, \ldots, e_T\}$. Each event is a tuple $e = \langle t, s, \sigma \rangle$, where $t \in \mathbb{N}^+$ is a unique timestamp, $s \in S$ is the sensor identifier, and $\sigma \in \{\text{ON}, \text{OFF}\}$ indicates the binary state change. The stream of events is ordered by the timestamp $t$.}

\added{Consistently with existing literature~\cite{arrotta2022dexar}, we discretize continuous sensor streams into virtual events. For instance, thanks to smart pre-processing steps, a sustained increase in humidity in the bathroom may generate a \textit{hot water running ON} event, while the subsequent value decrease may generate a \textit{hot water running OFF} event, as if these events were generated by a virtual sensor. Likewise, from the power signal of a smart plug,  multiple pairs of ON/OFF events can be extracted, corresponding to the electricity consumption pattern of specific appliances.}

\added{To process this continuous stream, a segmentation function divides $E$ into a set of (possibly overlapping) windows $\mathcal{W}$. 
Each window $W = \{e_1, \ldots, e_N\}$ is constructed using an \textbf{event-based} strategy (fixed $N$, ensuring uniform input size regardless of activity sparsity)}.


\subsection{\added{Feature Extractors and Encoder}}

\subsubsection{\added{Event-level Feature Extractor}}
\added{
Let $E$ be the stream of events as defined above. We first define the \emph{Event-level Feature Extractor} as a function:
$$
h_e : E \rightarrow \mathcal{Z}_e, \quad Z_e = h_e(e).
$$
that maps an event $e$ to a $d_e$-dimensional latent space $\mathcal{Z}_e$.
This representation captures the isolated semantic meaning of an event independently of its temporal context.} 

\subsubsection{\added{Contextualized Event-level Feature Extractor}}

\added{
In smart homes, context is critical. For instance, a motion sensor firing in the hallway has a fundamentally different meaning if preceded by a bedroom sensor rather than the front door. Therefore, we introduce a \emph{Contextualized Event-level Feature Extractor} to encode an event in the context of its specific window $W$. It is defined as a function
$$
h_{cxt} : E \times \mathcal{W} \rightarrow \mathcal{Z}_{cxt}, \quad Z_{cxt} = \Phi(h_e(e), \{h_e(e_i) : e_i \in W, e_i \neq e\})
$$
where $\Phi$ is a parametrized attention-based function operating on the embedded event space. By incorporating timestamps, $h_{cxt}$ encodes sequential ordering and relative temporal distances. Note that, for two different events in the same window, the contextualized event-level feature extractor produces a different representation 
$h_{cxt}(e_i,W) \neq h_{cxt}(e_j, W),\quad e_i,e_j \in W, i \neq j$}

\subsubsection{\added{Window Encoder}}


\label{subsubsec:transferability}
\added{
Given a window $W \in \mathcal{W}$, the \emph{Window Encoder} $g_\theta$ is a function that maps $W$ into the sequence of contextualized embeddings for all events within $W$:
$$
g_\theta: \mathcal{W} \to (\mathcal{Z}_{cxt})^N, \quad g_\theta(W) = \{h_{cxt}(e_1, W), \ldots, h_{cxt}(e_N, W)\}
$$
where $\theta$ denotes the learnable parameters in $h_e$ and $h_{cxt}$, which are optimized across diverse pre-training tasks to capture generalized behavioral patterns.}



\subsection{{\added{Window Encoder as a Foundation Model}}}
\label{win_encoder_as_foundation_model}
\added{
The function $g_\theta$ can be seamlessly composed with a task-specific function $f_\tau$, $\tau \in \mathcal{T}_{\text{d-stream}}$, to yield a final function $f_\tau(g_\theta(x))$, for any downstream task $\tau$. This composition supports three standard transfer paradigms, distinguished by which parameters are optimized for a given task $\tau$: (1) \textbf{identity mapping}, where $f_\tau$ is the identity function and $g_\theta$ is used directly, e.g.\ for unsupervised tasks; 
(2) \textbf{linear probing}, where only the parameters of $f_\tau$ are optimized while $\theta$ is held fixed, to prevent overfitting on small datasets; and (3) \textbf{full fine-tuning}, where both $\theta$ and the parameters of $f_\tau$ are jointly optimized for maximal task-specific performance.}

\added{In common machine learning terminology, $g_\theta$ plays the role of a pre-trained backbone and $f_\tau$ of a task-specific head, with linear probing and full fine-tuning corresponding to training the head alone or jointly with the backbone, respectively.}

\added{Since $g_\theta$ is adaptable to any downstream task, it can be considered a foundation model, according to the definition in \ref{def:foundation_model}, if it is trained on a sufficiently large and diverse dataset.
}

\section{\acronym{} Architecture and Training}
\begin{figure}
    \centering
    \includegraphics[width=0.8\linewidth]{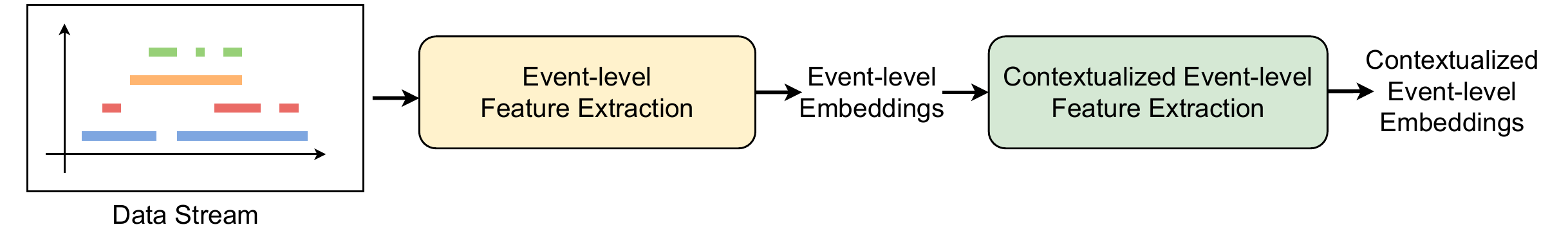}
    \captionsetup{singlelinecheck=false}
    \caption{High level architecture}
    \label{fig:high_level_architecture}
\end{figure}

Having defined the concept of a Foundation Model for \added{Behavioral Monitoring in Smart-home}, in this section, we now describe how we implement \acronym{}. A high-level representation of the architecture is illustrated in Figure \ref{fig:high_level_architecture}. The model processes the global data stream of events collected from the smart-home environment. This data stream feeds into two sequential processing stages: Event-level Feature Extraction and Contextualized Event-level Feature Extraction. These stages sequentially transform the raw sensor events into two types of learned representations: Event-level Embeddings and Contextualized Event-level Embeddings. The Event-level Feature Extraction module processes each sensor event individually to capture its intrinsic characteristics. In contrast, the Contextualized Event-level Feature Extraction module takes windows of event-level representations as input, and it incorporates temporal context from surrounding events (i.e., preceding and subsequent events) to produce enriched contextualized representations.
\added{Together, the event-level and contextualized event-level feature extractors constitute the Window Encoder ($g_\theta$), which forms the backbone of \acronym{} and is subsequently pretrained as the foundation model.}

\subsection{Event-level Feature Extraction}
\begin{figure}
    \centering
    \includegraphics[width=\linewidth]{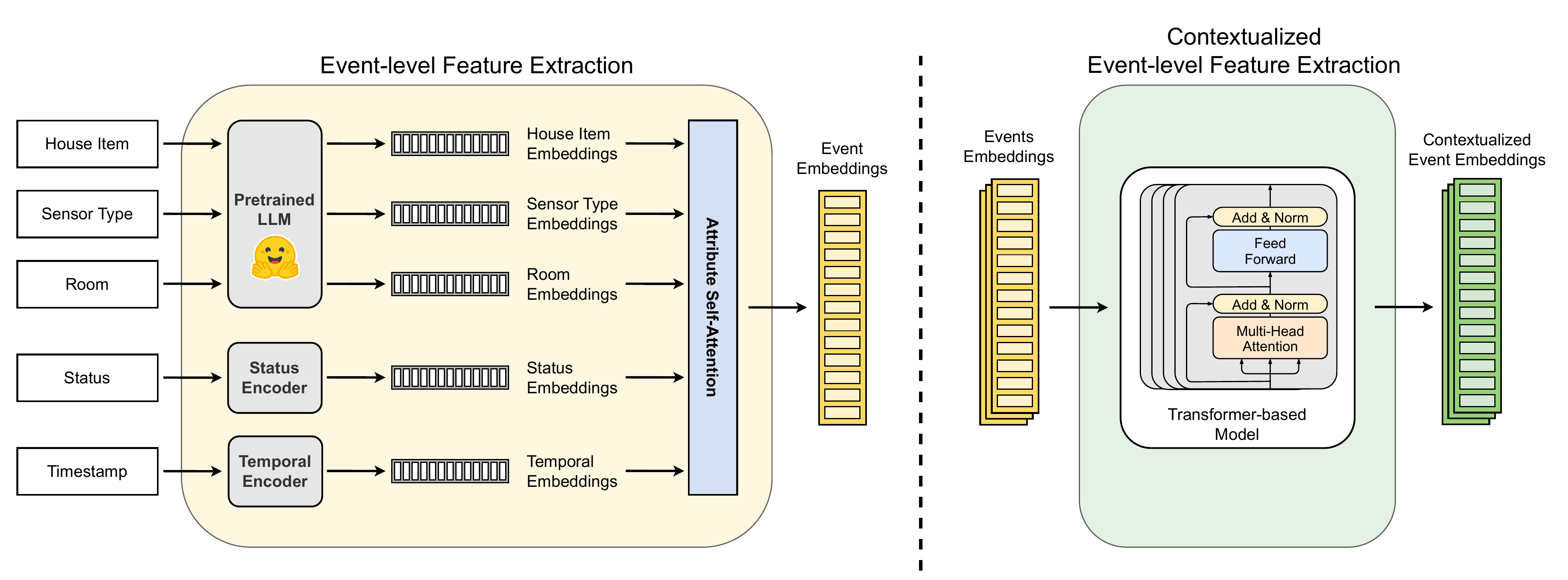}
    \vspace{0.5em}
    \begin{minipage}{0.59\linewidth}
        \centering
        \small\textsf{(a)\\Event-level Feature Extraction}
        \phantomsubcaption\label{fig:architecture_event_level}
    \end{minipage}
    \hfill
    \begin{minipage}{0.4\linewidth}
        \centering
        \small\textsf{(b)\\Contextualized Event-level Feature Extraction}        \phantomsubcaption\label{fig:architecture_contextualized_event_level}
    \end{minipage}
    \caption{Detailed architecture of \acronym's two main components. The Event-level Feature Extraction encodes sensor event attributes through specialized encoders and integrates them via Attribute Self-Attention. The Contextualized Event-level Feature Extraction uses transformer layers to capture temporal dependencies and produce context-aware event representations.}
\label{fig:architecture_detailed}
    \label{fig:architecture}
\end{figure}

The Event-level Feature Extraction module processes individual sensor events by encoding their various attributes into a unified representation, as shown in Figure \ref{fig:architecture_event_level}. Each sensor event is characterized by multiple attributes, including the house item, sensor type, room, status, and timestamp. These attributes are processed through three specialized encoders: a Pretrained LLM for semantic features, a Status Encoder for binary states, and a Temporal Encoder for time-related information. The resulting attribute-specific embeddings are then combined through an Attribute Self-Attention mechanism to produce the final event-level representation.

\subsubsection{Sensor Encoding}
As described earlier, each sensor is defined by three semantic attributes: the associated house item, the room in which it is located, and the sensor type. These attributes convey the core semantic meaning of a sensor event and are therefore critical for accurate modeling. To effectively capture this semantic information, we leverage a pretrained LLM, which provides rich representations grounded in prior knowledge learned from large-scale textual data. Specifically, each attribute is represented by its textual name, processed independently, and mapped to a dense vector in the pretrained LLM’s embedding space. 
This approach allows the sensor attributes to be embedded in a semantically meaningful space that enhances generalization. For example, semantically related appliances such as a stove and an oven are naturally embedded closer together than unrelated ones like a stove and a bed, capturing domain-relevant relationships that traditional encoding schemes, such as one-hot encoding, would not represent.

\subsubsection{Status Encoding}
The status attribute in our model is inherently simple, as it can only assume two primary values: ON or OFF (with an additional MASK state considered during pretraining). Unlike other attributes that may require richer representations, the status information does not necessitate complex encoding. Therefore, we represent the status using a standard embedding approach that maps each possible state to a learned dense vector in the model’s hidden space. 
The semantic interpretation of these states is context-dependent: for instance, ON for a motion sensor indicates presence, while ON for a magnetic switch indicates that an object has been opened.

\subsubsection{Temporal Encoding}
To capture temporal patterns in the data, we extract three time-related attributes from the timestamp: day of the week, hour of the day, and seconds within the hour. Coarser temporal attributes, such as the month, are deliberately excluded, as most datasets span only one or a limited number of months, which would be insufficient for learning meaningful periodic behaviors or habits. The selected attributes reflect recurring daily and weekly cycles that are more consistently observable across datasets.

Different encoding strategies are employed depending on the nature of each temporal attribute. For the day of the week and the hour of the day, we adopt a cyclical encoding that explicitly models their periodic structure \cite{mardia2009directional}. Each value is transformed using multiple harmonic frequencies of the corresponding base period and represented as a sequence of sine–cosine pairs. This formulation allows the model to preserve temporal continuity and cyclic proximity (e.g., between consecutive hours or days). In line with prior work \cite{lee2025adaptive, liu2025marauder}, a subsequent projection is applied to both representations, enabling the model to learn which frequency components are most informative, thereby improving robustness to sparse or irregular temporal sampling. In contrast, the seconds-within-the-hour attribute is treated as a discrete variable and encoded using a learned embedding representation, as it does not exhibit meaningful cyclical continuity. Unlike hours or days, where boundary values are semantically close (e.g., 23:59 and 00:01 both correspond to midnight), the wrap-around of seconds coincides with an hour transition, representing a discontinuity rather than a natural cycle.

\subsubsection{Attribute Self-Attention}
The encoded attributes are aggregated into a unified event-level representation through a self-attention mechanism applied across all previously described attribute dimensions. This choice is motivated by the observation that individual attributes do not carry meaning in isolation, but rather in relation to each other: for instance, a kitchen sensor activation at 7:00 AM suggests a different activity than the same activation at 11:00 PM. The self-attention mechanism explicitly models these inter-attribute dependencies, producing a unified representation that captures their contextual interactions. As a result, each event is represented by a single vector that serves as the foundational unit for subsequent modeling.

\subsection{Contextualized Event-level Feature Extraction}
The Contextualized Event-level Feature Extraction module enriches the individual event representations by incorporating information from surrounding events in the window, as illustrated in Figure \ref{fig:architecture_contextualized_event_level}. 
This module takes as input the event embeddings of all the events in a window and processes them sequentially through a transformer-based architecture. The core of this module consists of stacked transformer encoder layers. 
The self-attention mechanism enables each event to attend to all other events in the window, allowing the model to capture sequential dependencies, temporal patterns, and contextual relationships between events. This design allows the model to learn complex interactions between events and produce Contextualized Event Embeddings that encode not only the intrinsic characteristics of each event but also its role and significance within the broader window.

\subsection{Pretraining \acronym{}}
\label{sec:pretrain}

\begin{figure}
    \centering
    \includegraphics[width=1\linewidth]{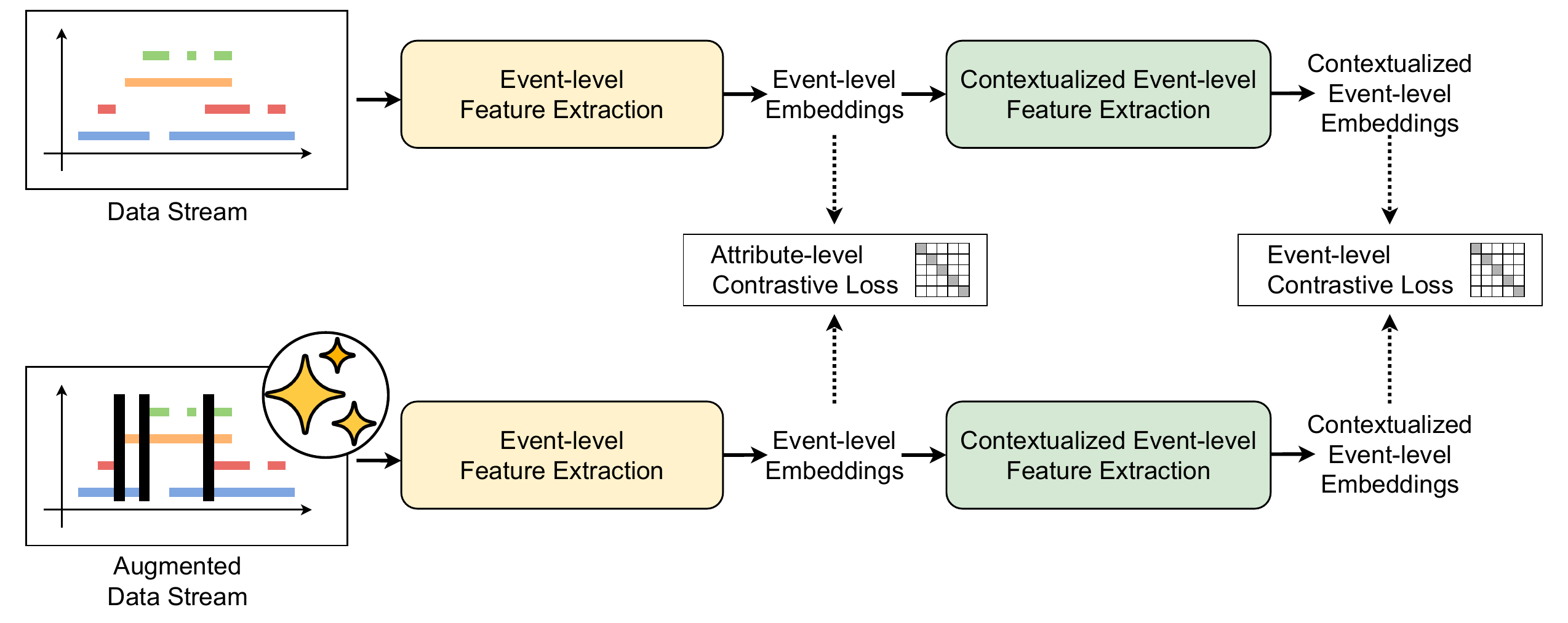}
    \caption{Dual contrastive learning framework for pretraining \acronym{}. Attribute-level contrastive loss learns robust event representations by masking individual attributes, while event-level contrastive loss captures temporal context by masking entire events. Both stages use augmented data streams to learn representations through contrastive objectives.}
\label{fig:pretraining_framework}
    \label{fig:overall_pretrain}
\end{figure}

To pretrain our model, we adopted a self-supervised dual pretraining paradigm inspired by \cite{li2023urban} and depicted in Figure \ref{fig:overall_pretrain}. The first phase of this dual contrastive learning framework employs an \textbf{attribute-level} contrastive loss. In this phase, we augment events by masking one attribute at a time (i.e., substituting the value of that attribute with a "MASK" token). Events are randomly selected for augmentation according to a predefined probability. For each selected event, we randomly choose which specific attribute to mask. Both the original and the augmented sequences are then independently processed by the model to obtain their corresponding sequence-level embeddings. The original sequence and its masked augmentation form a positive pair, while embeddings from other sequences within the same batch are treated as negative samples. The model is then trained using an InfoNCE loss \cite{oord2018representation} to learn robust attribute-level representations. 


The second phase introduces an \textbf{event-level} contrastive loss, where we augment entire sequences by masking complete events with a certain probability, applying the masking operation to all attributes of the selected events simultaneously (in this case, we substitute the value of all the attributes of that event with a "MASK" token). Similarly, the original sequence and its masked augmentation form a positive pair, while embeddings from other sequences within the same batch are treated as negative samples. This phase also employs an InfoNCE loss for optimization. Following the dual contrastive learning paradigm, the event-level feature extractor is frozen after the first phase, ensuring that the model builds upon the learned attribute-level representations while developing higher-level contextual understanding during the second phase.

\subsection{Task-specific Heads}
\label{subsec:task_specific_heads}
The components described so far constitute the general-purpose feature extractor that produces rich, contextualized representations of sensor event sequences. To adapt these representations to a specific downstream task, we append a task-specific head on top of the shared encoder during fine-tuning. The task-specific head is designed according to the task of interest and operates on the contextualized event-level embeddings produced by the encoder. This modular design enables the reuse of the same feature extraction backbone across multiple tasks while allowing flexible and targeted optimization for each application. The specific heads we employed in this paper are explained, for each downstream task, in section \ref{sec:experimental_evaluation}. It is important to note that the primary contribution of this work lies in the feature extraction framework itself; the task-specific heads employ standard architectures and are included to demonstrate the applicability and effectiveness of the learned representations. 


\section{Considered Smart-home Datasets}
\label{sec:datasets}
In this section, we provide an overview of the datasets considered in this work for both pretraining and downstream tasks. We selected publicly available datasets, ensuring diversity in sensors, environments, users, and activities. It is important to note that, as described in Section \ref{sec:experimental_setup}, when a dataset is used for fine-tuning and testing on downstream tasks, no part of that dataset is included in the pretraining phase, ensuring a proper evaluation of the model's generalization capabilities.

\subsection{Datasets Descriptions}
\subsubsection{CASAS Datasets}
The CASAS smart-home project provides a comprehensive repository of annotated sensor data for activity recognition research \cite{cook2013casas}. The datasets are collected from real smart homes instrumented with PIR motion sensors, magnetic sensors, and temperature sensors. While the CASAS repository include a high number of different datasets, we focused on the two widely adopted Aruba and Milan. That's because they both capture single-resident scenarios, and are commonly employed as standard benchmarks. Moreover, including additional CASAS datasets would create a severely unbalanced evaluation that could bias our models toward CASAS-specific characteristics.

\subsubsection{Van Kasteren Datasets}
The van Kasteren datasets represent one of the earliest and most influential benchmark collections for ADLs in smart homes \cite{van2011human}. The datasets were collected from real-world residences equipped with wireless sensor networks using simple, non-intrusive binary sensors, including reed switches, pressure mats, PIR motion detectors, mercury contacts, and float sensors. We utilized two houses from this collection: House A and House C. We excluded House B from our experiments due to the lack of explicit semantic information about the sensors, which is crucial for our approach.

\subsubsection{UCI Datasets}
The UCI ADL Binary datasets, introduced by Ordóñez et al. \cite{ordonez2013activity}, comprise data collected from two real smart homes (A and B) instrumented with binary sensors, including PIR motion sensors, pressure sensors, magnetic sensors (reed switches), float sensors, and plug sensors for electrical appliances. We excluded Home A from our experiments as previous works have demonstrated that this dataset is too simple and does not provide additional informative value for evaluation \cite{civitarese2025large}.


\subsubsection{Orange Datasets}
The Orange Labs datasets comprise two collections for ambient sensor-based activity recognition. Orange4Home \cite{cumin2017dataset} includes 236 heterogeneous sensors, including binary sensors, water consumption meters, power monitors, and environmental sensors (luminosity, CO2, noise) from a single occupant. MuRAL (Multi-Resident Ambient sensor dataset with natural Language) \cite{chen2025mural} extends to multi-resident scenarios with 23 sensors (PIR motion sensors, magnetic contacts, and smart plugs), capturing sessions with 2 to 4 participants. Although from the same provider, these datasets were collected with a significant temporal gap using completely different sensor configurations and experimental scenarios, making them complementary rather than redundant.

\section{Experimental Evaluation}
\label{sec:experimental_evaluation}
In this section, we demonstrate how \acronym{} aligns with the core properties of a foundation model by assessing its adaptability and generalization capabilities across multiple datasets and tasks, achieving consistent improvements over state-of-the-art baselines even under severe data scarcity. 

\subsection{Experimental Setup}
\label{sec:experimental_setup}
To comprehensively evaluate \acronym{} and demonstrate its ability to generalize to previously unseen environments, we employed a rigorous leave-one-dataset-out evaluation \cite{presotto2023combining} across all seven datasets described in Section \ref{sec:datasets}. In this evaluation strategy, six datasets were used exclusively for pretraining, while the seventh dataset was held out entirely and used only for fine-tuning and testing on downstream tasks. This means that no data from the held-out dataset was exposed to the model during pretraining. The fine-tuning and test data represent completely unseen smart-home environments and users. This procedure was repeated seven times, with each dataset serving as the held-out evaluation dataset exactly once, ensuring a thorough assessment of the model's cross-dataset generalization capabilities.
Note that, in all stages (pretraining, fine-tuning, and testing), we include continuous sensor data, like temperature readings, converted into virtual binary events during preprocessing (see Section~\ref{subsec:sensor_event_stream_and_segmentation}).

\subsubsection{\added{Handling the "Other" Class}}
\label{subsubsec:other_class}
\added{All datasets considered in this work include a class named "Other", representing transition periods between activities or time intervals in which the subject was performing an activity not explicitly considered within the dataset's annotation scheme. This class is particularly challenging, as it typically introduces noise and ambiguity that can mislead learned models, often resulting in degraded performance. As a consequence, the majority of existing works in the literature opt to exclude windows labeled as "Other" from their experiments. In contrast, we deliberately retain all such windows across every stage of our pipeline, from pretraining to fine-tuning and evaluation on downstream tasks, treating the class "Other" on equal footing with all other activity classes. This choice reflects a commitment to more realistic experimental conditions, as discarding such windows would artificially simplify the problem and obscure the true difficulty of deploying activity recognition models in real smart-home environments.}

\subsubsection{Pretraining}
For the pretraining phase, we utilized all available data from each of the six training datasets in their entirety, following the dual contrastive learning framework described in Section \ref{sec:pretrain}. To address the substantial size imbalances across datasets (where some datasets contain significantly more samples than others), we applied random oversampling at the dataset level. This technique ensures that smaller datasets contribute proportionally during training, preventing the model from being disproportionately influenced by larger datasets and promoting balanced representation learning across diverse smart-home environments.

\subsubsection{Fine-tuning and testing for Downstream Tasks}
\label{subsubsec:fine_tuning_testing}
To simulate realistic data scarcity scenarios commonly encountered in practical smart-home deployments, we introduced a parameter called the Training Data \%, representing the percentage of available labeled training data from the held-out dataset. For each Data scarcity scenario, we randomly sampled the corresponding percentage of instances from the unseen downstream dataset and conducted fine-tuning exclusively on this limited subset. \added{Note that, following strategy (3) of Section \ref{win_encoder_as_foundation_model}, we fine-tune both the Window Representation and the task-specific head jointly. Given the \acronym{}'s small size (36M parameters), full fine-tuning remains computationally lightweight even in low-data regimes.}

We systematically evaluated our model across four Training Data percentages: 5\%, 10\%, 15\%, and 30\% of the training data. The robustness and statistical reliability of our results were ensured through a standard 5-fold cross-validation applied during the fine-tuning and evaluation phase, with performance metrics averaged across all folds.
For the fine-tuning phase on downstream tasks, we trained for 10 epochs. Notably, we did not employ early stopping during fine-tuning because the extremely limited amount of labeled training data made it impractical to extract a separate validation set without significantly compromising model performance. Instead, we relied on cross-validation across folds to ensure robust evaluation.

\subsection{Implementation Details} 
\subsubsection{\acronym's Configuration}
For the pretrained LLM component used in sensor encoding, we employed a Sentence-BERT model, specifically the \texttt{all-MiniLM-L6-v2} variant from the sentence-transformers library \footnote{\url{https://huggingface.co/sentence-transformers}}. We fixed the dimensionality $d$ of the latent space at $384$ for both event-level and contextualized event-level representations. 
For the multi-head attention mechanism in the contextualized event-level feature extraction module, we set the number of attention heads to 12 and the number of transformer layers to 12, following standard practices established in the transformer literature. These architectural choices were designed to strike an appropriate balance between model expressiveness and optimization stability. Unlike domains such as natural language processing, where foundation models are pretrained on massive corpora containing billions of tokens, the smart-home domain, even when aggregating multiple publicly available datasets, provides substantially less pretraining data. Consequently, employing an excessively large model with billions of parameters would risk overfitting and poor generalization. 
Our implementation contains a total of $36,075,240$ parameters, of which approximately $20$ million belong to the pretrained Sentence-BERT model and are kept frozen during training, while the remaining ones are trainable. We found this model size to be well-suited for the scale of available smart-home data. The complete framework was implemented in Python 3 using the PyTorch library for model development. \footnote{We published the code of \acronym{} as open-source software, available at \url{https://gitlab.di.unimi.it/ewlab/public/domusfm}}


\subsubsection{Segmentation} For processing the continuous stream of sensor events, we employed a fixed event-based segmentation strategy, setting the window size to $N=30$ consistently with the literature \cite{krishnan2014activity, fiori2026improving}. To ensure comprehensive coverage of the entire data stream, we used an overlap of $29$ events between consecutive windows, resulting in a sliding window that advances by one event at a time.

\subsection{\added{Baselines}}
\added{
We consider a baseline for the backbone and an end-to-end baseline for each specific downstream task.
As a backbone baseline, in the absence of an event-native foundation model for time series, we adopt Chronos-2, a well-known transformer-based foundation model for time-series forecasting developed by Amazon}\footnote{\url{https://huggingface.co/amazon/Chronos-2}}.
\added{The representations produced by Chronos-2 are then fed into the same task-specific head as for our backbone. As explained in Section \ref{subsubsec:fine_tuning_testing}, we adopt a full fine-tuning strategy for our model. To ensure a fair comparison, we apply the same strategy to this baseline, fine-tuning the entire Chronos-2 backbone together with the task-specific head for each downstream task.}

\added{For each downstream task, we also consider an end-to-end task-specific baseline, which is described together with the respective experimental setup in the following subsections. As done in the referenced literature, these end-to-end baselines are trained only on task-specific data.}

\subsection{Downstream task 1: ADL Recognition}
\subsubsection{Problem Formulation}
Before presenting our experimental results, we formally define the ADL recognition task addressed in this work. 
Let $W \in \mathcal{W} $ be a window of $N$ consecutive sensor events, and let $\mathcal{A}=\{a_1,a_2,...,a_M\}$ be the set of $M$ target activities of daily living of interest (e.g., cooking, sleeping, grooming). 
\added{As discussed in Section \ref{subsubsec:other_class}, we include the \emph{Other} activity class. For the purposes of training and evaluation, this class is treated in the same manner as all other activity classes.}
The ADL recognition problem consists of determining the activity $a \in \mathcal{A}$ that was being performed (or initiated) at the time of the last event $e_N$ in the window $W$. In other words, given a sequence of sensor events leading up to and including a specific moment in time, the task is to classify which activity the resident was engaged in at that moment.
It is important to note that this is not the only possible formulation used in the human activity recognition domain. The formulation we adopt follows established works in the literature \cite{krishnan2014activity, fiori2026improving} and is particularly well-suited for real-time activity monitoring applications, where the system must continuously infer the current activity based on the most recent sensor observations. This formulation also aligns with the sliding window approach commonly employed in smart-home activity recognition systems.

\subsubsection{\added{Task-Specific} Baseline}
For comparison, we employed DeepCASAS \cite{liciotti2020sequential} as our \added{end-to-end task-specific} baseline, which represents a widely-adopted \added{fully supervised} deep learning baseline for activity recognition in smart-home environments. DeepCASAS utilizes a bidirectional LSTM network architecture that captures both forward and backward temporal dependencies in sensor event streams. We implemented the model using the identical architecture and hyperparameters as specified in the original paper, ensuring a fair comparison. 

\subsubsection{ADL Recognition Head}
For the ADL recognition downstream task, we implement the head as a simple linear layer that takes the contextualized event embeddings as input and outputs a probability distribution over the target classes.

\subsubsection{Results}
For performance assessment, we utilized standard classification metrics, namely precision, recall, and F1-score, with weighted F1-scores reported in the results tables. The detailed results are presented in Table \ref{tab:adls_recognition_f1_scores}, which displays performance across all datasets and Train Data percentages. \acronym{} consistently outperforms the DeepCASAS baseline across all seven datasets for all Train Data percentages. 
The improvements are particularly notable in highly constrained settings, where \acronym{} achieves significant gains over baseline performance, with the largest benefits seen on some of the more challenging datasets. Even when more labeled data is available, \acronym{} continues to outperform alternatives, demonstrating that it not only excels in few-shot scenarios but also leverages additional supervision effectively. These results highlight the strength of our foundation model approach, showing that pretraining on diverse smart-home environments enables robust transfer learning and strong performance even in minimally supervised target settings.

\begin{figure}
    \centering

    \begin{subfigure}{0.9\textwidth}
    \centering
    \includegraphics[width=\textwidth]{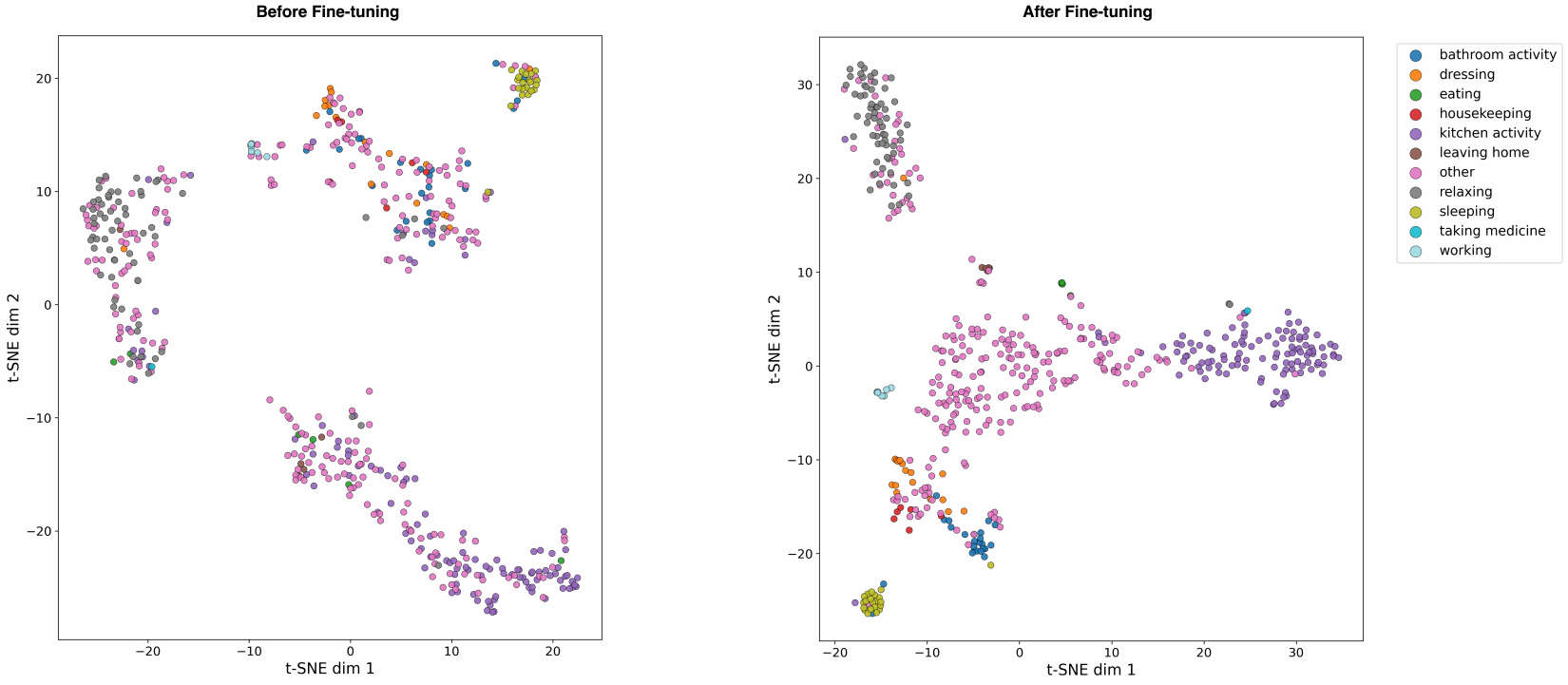}
    \caption{CASAS Milan}
    \end{subfigure}

    \vspace{1em}

    \begin{subfigure}{0.9\linewidth}
        \centering
        \includegraphics[width=\linewidth]{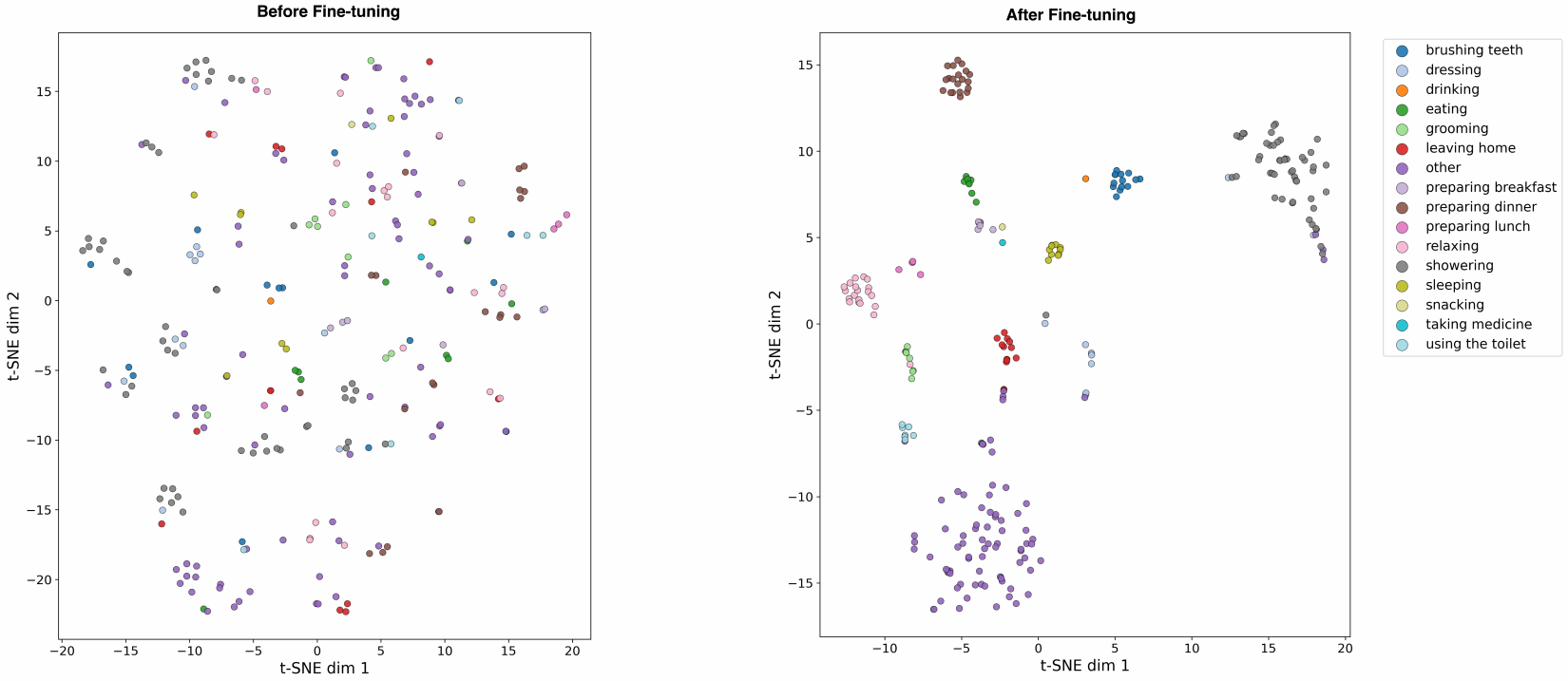}
        \caption{Kasteren C}
        \label{fig:embeddings_kasteren}
    \end{subfigure}

    \caption{\added{t-SNE visualization of the learned embeddings before
    (left) and after (right) fine-tuning, for CASAS Milan (top) and
    Kasteren C (bottom), using 5\% of the fine-tuning data. Points are
    colored by ground-truth class. Before fine-tuning, embeddings form
    clusters that do not align with class labels, reflecting the
    self-supervised nature of pretraining. After fine-tuning, clusters
    correspond closely to the classes, showing that our model effectively
    adapts to the downstream task even with limited fine-tuning data.}}
    \label{fig:embeddings}
\end{figure}

\added{Figure \ref{fig:embeddings_milan} and \ref{fig:embeddings_kasteren} shows the t-SNE visualization of the learned embeddings before and after fine-tuning, for two representative datasets, namely CASAS Milan and Kasteren C, using the smallest fine-tuning data percentage (5\%). We observe that after pretraining, some clusters already emerge, but they do not necessarily correspond to the class labels. This is expected, since pretraining is performed in a self-supervised manner, without access to class information, and therefore the resulting structure reflects general patterns in the underlying data rather than task-specific class boundaries. After fine-tuning, the embeddings reorganize into clusters that align clearly with the classes, demonstrating the effectiveness of our model in adapting the pretrained representations to the downstream classification task even with a very limited amount of fine-tuning data.
}

\begin{table}[h]
\centering
\footnotesize
\caption{ADL recognition performance (Weighted F1-score) comparing \acronym{} against the DeepCASAS and Chronos-2 baselines across seven smart-home datasets and different Train Data percentages (5\%, 10\%, 15\%, 30\%). Results are averaged over 5-fold cross-validation. \acronym{} outperforms the baseline for all Train Data percentages, across all datasets.}

\begin{tabular}{lcccccc}
\toprule
Train Data \% & \multicolumn{3}{c}{5\%} & \multicolumn{3}{c}{10\%} \\
\midrule
 & DeepCASAS & \added{Chronos-2} & \acronym & DeepCASAS & \added{Chronos-2} & \acronym \\
\midrule
CASAS Milan   & 0.54 & \added{0.54} & \textbf{0.72} & 0.55 & \added{0.56} & \textbf{0.82} \\
CASAS Aruba   & 0.71 & \added{0.70} & \textbf{0.76} & 0.72 & \added{0.72} & \textbf{0.78} \\
Kasteren A    & 0.25 & \added{0.21} & \textbf{0.48} & 0.34 & \added{0.31} & \textbf{0.56} \\
Kasteren C    & 0.31 & \added{0.30} & \textbf{0.59} & 0.35 & \added{0.35} & \textbf{0.67} \\
Orange4Home   & 0.77 & \added{0.79} & \textbf{0.89} & 0.80 & \added{0.82} & \textbf{0.91} \\
MuRal         & 0.21 & \added{0.24} & \textbf{0.60} & 0.26 & \added{0.29} & \textbf{0.70} \\
UCI B         & 0.21 & \added{0.20} & \textbf{0.38} & 0.22 & \added{0.23} & \textbf{0.46} \\
\bottomrule
\end{tabular}

\vspace{1em} 

\begin{tabular}{lcccccc}
\toprule
Train Data \% & \multicolumn{3}{c}{15\%} & \multicolumn{3}{c}{30\%} \\
\midrule
 & DeepCASAS & \added{Chronos-2} & \acronym & DeepCASAS & \added{Chronos-2} & \acronym \\
\midrule
CASAS Milan   & 0.56 & \added{0.58} & \textbf{0.86} & 0.57 & \added{0.60} & \textbf{0.92} \\
CASAS Aruba   & 0.72 & \added{0.74} & \textbf{0.82} & 0.73 & \added{0.76} & \textbf{0.91} \\
Kasteren A    & 0.42 & \added{0.45} & \textbf{0.61} & 0.47 & \added{0.49} & \textbf{0.68} \\
Kasteren C    & 0.36 & \added{0.48} & \textbf{0.72} & 0.40 & \added{0.53} & \textbf{0.81} \\
Orange4Home   & 0.81 & \added{0.83} & \textbf{0.94} & 0.84 & \added{0.85} & \textbf{0.96} \\
MuRal         & 0.29 & \added{0.31} & \textbf{0.71} & 0.34 & \added{0.37} & \textbf{0.80} \\
UCI B         & 0.30 & \added{0.26} & \textbf{0.51} & 0.38 & \added{0.39} & \textbf{0.60} \\
\bottomrule
\end{tabular}

\label{tab:adls_recognition_f1_scores}
\end{table}

\subsection{Downstream task 2: Next-\textit{k} Events Prediction}

\subsubsection{Problem Formulation}
The next-\textit{k} events prediction task addresses a different aspect of smart-home understanding compared to ADL recognition. Given a window $W \in \mathcal{W}$ of $N$ consecutive sensor events and the set of sensors $S$ deployed in the environment, the task consists of predicting the next-\textit{k} events that will occur in the event stream. Formally, for each sensor $s \in S$ and each status $\sigma \in \{\text{ON}, \text{OFF}\}$, we predict the count $C_{(s,\sigma)}$ representing how many times the event $(s,\sigma)$ will occur in the next-\textit{k} events. 

Our formulation diverges from traditional sequence prediction tasks.
We are not interested in predicting the exact temporal order in which events will occur, nor their precise timestamps. Instead, we adopt a bag-of-events approach, focusing on predicting which sensor events will occur, with what status (ON or OFF), and how many times each event type will appear within the next-\textit{k} events. 
\added{This formulation is justified by the 
interest in predicting patterns of events that characterise a behavior independently from their fine-grained internal order of events.}
%
%
%
For instance, when a user is preparing a drink, whether they first open the refrigerator to retrieve water and then open the glass cabinet to get a glass, or perform these actions in reverse order, represents negligible behavioral variation at this granularity. Both sequences are part of the same atomic activity and are functionally equivalent from a behavioral modeling perspective. 
\added{Similarly, when a user is taking medication, 
whether they first retrieve the medicine and then pour a glass of water, or vice versa, constitutes the same high-level behavior.}
\added{Even in other application domains like energy management, proactive home automation, or anomaly detection, it is useful to understand which resources will be accessed and with what frequency in the near future, rather than their precise microsecond-level ordering within an ongoing activity.}

\added{Predicting the exact order of events can be considered an additional downstream task with other applications. For example, at a coarser granularity, we may want to predict whether a user will cook and then eat versus eat and then cook, or we may want to identify unusual ordering of atomic actions to perform an activity of daily living. This, as well as many other downstream tasks, is worth considering in future work.}


For our event prediction task, we evaluated \acronym{} on two prediction horizon $k = 10$ and $k=30$ events. The $k=30$ aligns with the window's length, while with $k=10$ we test the performance over a shorter, more immediate period. Comparing these horizons reveals how accuracy changes as the forecasting window varies.

\subsubsection{\added{Task-Specific} Baseline}
To the best of our knowledge, no existing approach in the literature specifically addresses next-\textit{k} events prediction using our bag-of-events formulation. Therefore, we adapted the method proposed by Takeda et al. \cite{takeda2024sensor}, which represents the most recent work on events prediction in smart homes and is the closest to our formulation. The original approach employs a generative transformer architecture inspired by the GPT-2 model for predicting sequences of future sensor events, trained from scratch with smart-home sensor event sequences. We modified the inference procedure to predict a fixed number of \textit{k} events following our bag-of-events approach, where the model outputs which events will occur and their respective counts. Apart from this adaptation to match our problem formulation, we trained the baseline using the same architecture and hyperparameters specified in the original paper to ensure a fair comparison.

\subsubsection{Next-\textit{k} Events Prediction Head}
For next-\textit{k} events prediction, we implement the head as a dual-head architecture consisting of two separate linear layers: the first predicts a distribution over possible future event types, while the second predicts the expected count for each event type within the next-\textit{k} events.

\subsubsection{Results}
For performance assessment, we adopted an adaptation of precision and recall metrics specifically designed for bag-of-events (multisets). In particular:
Let $ M_{\text{GT}}$ be the multiset of events in the ground truth sequence, $M_{\text{PRED}}$ the predicted multiset of events, we can define
$$
\text{Precision} = \frac{|M_{\text{GT}} \cap M_{\text{PRED}}|}{|M_{\text{PRED}}|}, \quad
\text{Recall} = \frac{|M_{\text{GT}} \cap M_{\text{PRED}}|}{|M_{\text{GT}}|}, \quad
\text{F1} = \frac{2 \cdot \text{Precision} \cdot \text{Recall}}{\text{Precision} + \text{Recall}}
$$
The results are therefore expressed in terms of F1-score. Since we are evaluating multiset predictions rather than individual class assignments, only an overall F1-score is computed, with no distinction between macro or weighted variants. Detailed results are presented in Table \ref{tab:next-30 prediction} for $k = 30$ and Table \ref{tab:next-10 prediction} for $k = 10$ across all datasets. Consistent with the ADL recognition task, our model substantially outperforms the baseline across all Train Data percentages in both cases, further validating the robustness and effectiveness of our approach under varying degrees of data scarcity.

\begin{table}[h]
\centering
\footnotesize
\caption{Next-$30$ events prediction performance (F1-score) comparing \acronym{} against the GPT-2-based and Chronos-2 baselines across seven smart-home datasets and different Train Data percentages (5\%, 10\%, 15\%, 30\%). Results are averaged over 5-fold cross-validation. \acronym{} outperforms the baseline in all scenarios across all datasets.}

\begin{tabular}{lcccccc}
\toprule
Train Data \% & \multicolumn{3}{c}{5\%} & \multicolumn{3}{c}{10\%} \\
\midrule
 & GPT-2 & \added{Chronos-2} & \acronym & GPT-2 & \added{Chronos-2} & \acronym \\
\midrule
CASAS Milan  & 0.39 & \added{0.42} & \textbf{0.57} & 0.49 & \added{0.51} & \textbf{0.61} \\
CASAS Aruba  & 0.52 & \added{0.55} & \textbf{0.59} & 0.55 & \added{0.55} & \textbf{0.62} \\
Kasteren A   & 0.29 & \added{0.39} & \textbf{0.66} & 0.29 & \added{0.41} & \textbf{0.74} \\
Kasteren C   & 0.46 & \added{0.48} & \textbf{0.59} & 0.47 & \added{0.50} & \textbf{0.69} \\
Orange4Home  & 0.70 & \added{0.73} & \textbf{0.86} & 0.74 & \added{0.79} & \textbf{0.87} \\
MuRal        & 0.63 & \added{0.64} & \textbf{0.69} & 0.65 & \added{0.67} & \textbf{0.75} \\
UCI B        & 0.36 & \added{0.48} & \textbf{0.76} & 0.54 & \added{0.59} & \textbf{0.82} \\
\bottomrule
\end{tabular}

\vspace{1em}

\begin{tabular}{lcccccc}
\toprule
Train Data \% & \multicolumn{3}{c}{15\%} & \multicolumn{3}{c}{30\%} \\
\midrule
 & GPT-2 & \added{Chronos-2} & \acronym & GPT-2 & \added{Chronos-2} & \acronym \\
\midrule
CASAS Milan  & 0.55 & \added{0.59} & \textbf{0.65} & 0.61 & \added{0.65} & \textbf{0.73} \\
CASAS Aruba  & 0.60 & \added{0.59} & \textbf{0.65} & 0.63 & \added{0.63} & \textbf{0.70} \\
Kasteren A   & 0.49 & \added{0.57} & \textbf{0.80} & 0.64 & \added{0.71} & \textbf{0.87} \\
Kasteren C   & 0.47 & \added{0.55} & \textbf{0.75} & 0.67 & \added{0.71} & \textbf{0.86} \\
Orange4Home  & 0.79 & \added{0.81} & \textbf{0.89} & 0.82 & \added{0.85} & \textbf{0.89} \\
MuRal        & 0.73 & \added{0.75} & \textbf{0.79} & 0.76 & \added{0.79} & \textbf{0.84} \\
UCI B        & 0.60 & \added{0.67} & \textbf{0.86} & 0.76 & \added{0.81} & \textbf{0.90} \\
\bottomrule
\end{tabular}

\label{tab:next-30 prediction}
\end{table}

\begin{table}[h]
\centering
\footnotesize
\caption{Next-$10$ events prediction performance (F1-score) comparing \acronym{} against the GPT-2-based and Chronos-2 baselines across seven smart-home datasets and different Train Data percentages (5\%, 10\%, 15\%, 30\%). Results are averaged over 5-fold cross-validation. \acronym{} outperforms the baseline in all scenarios across all datasets.}

\begin{tabular}{lcccccc}
\toprule
Train Data \% & \multicolumn{3}{c}{5\%} & \multicolumn{3}{c}{10\%} \\
\midrule
 & GPT-2 & \added{Chronos-2} & \acronym & GPT-2 & \added{Chronos-2} & \acronym \\
\midrule
CASAS Milan  & 0.37 & \added{0.39} & \textbf{0.52} & 0.41 & \added{0.44} & \textbf{0.56} \\
CASAS Aruba  & 0.47 & \added{0.48} & \textbf{0.54} & 0.50 & \added{0.49} & \textbf{0.57} \\
Kasteren A   & 0.28 & \added{0.28} & \textbf{0.45} & 0.43 & \added{0.45} & \textbf{0.54} \\
Kasteren C   & 0.30 & \added{0.29} & \textbf{0.52} & 0.28 & \added{0.36} & \textbf{0.61} \\
Orange4Home  & 0.62 & \added{0.65} & \textbf{0.79} & 0.64 & \added{0.69} & \textbf{0.83} \\
MuRal        & 0.45 & \added{0.54} & \textbf{0.68} & 0.56 & \added{0.59} & \textbf{0.71} \\
UCI B        & 0.32 & \added{0.39} & \textbf{0.62} & 0.42 & \added{0.45} & \textbf{0.68} \\
\bottomrule
\end{tabular}

\vspace{1em}

\begin{tabular}{lcccccc}
\toprule
Train Data \% & \multicolumn{3}{c}{15\%} & \multicolumn{3}{c}{30\%} \\
\midrule
 & GPT-2 & \added{Chronos-2} & \acronym & GPT-2 & \added{Chronos-2} & \acronym \\
\midrule
CASAS Milan  & 0.51 & \added{0.53} & \textbf{0.58} & 0.52 & \added{0.56} & \textbf{0.62} \\
CASAS Aruba  & 0.50 & \added{0.52} & \textbf{0.59} & 0.52 & \added{0.54} & \textbf{0.60} \\
Kasteren A   & 0.53 & \added{0.52} & \textbf{0.58} & 0.62 & \added{0.59} & \textbf{0.70} \\
Kasteren C   & 0.30 & \added{0.39} & \textbf{0.66} & 0.39 & \added{0.45} & \textbf{0.75} \\
Orange4Home  & 0.75 & \added{0.72} & \textbf{0.85} & 0.79 & \added{0.79} & \textbf{0.85} \\
MuRal        & 0.58 & \added{0.61} & \textbf{0.72} & 0.67 & \added{0.70} & \textbf{0.76} \\
UCI B        & 0.46 & \added{0.50} & \textbf{0.70} & 0.51 & \added{0.59} & \textbf{0.76} \\
\bottomrule
\end{tabular}

\label{tab:next-10 prediction}
\end{table}

\subsection{\added{Downstream task 3: Unsupervised Clustering}}
\label{subsec:unsupervised_clustering}

\subsubsection{\added{Problem Formulation}}

\added{
The unsupervised clustering task explores a fundamentally different paradigm compared to the previous downstream tasks: rather than optimizing for a predefined prediction objective, it aims to discover latent structure in smart-home sensor data without any supervision signal. Let $W \in \mathcal{W}$ be a window of $N$ consecutive sensor events, and let $\mathbf{z}_W \in \mathbb{R}^d$ denote the $d$-dimensional representation extracted by \acronym{} for that window. Given a collection of such representations $\mathcal{Z} = \{\mathbf{z}_{W_1}, \mathbf{z}_{W_2}, \ldots, \mathbf{z}_{W_n}\}$, the clustering task consists of partitioning $\mathcal{Z}$ into $K$ groups such that representations corresponding to behaviorally similar windows are assigned to the same cluster. No activity labels or any other form of supervision are used at any stage of this process, neither during the extraction of representations nor during clustering assignment. 
The motivation for this task is to assess whether the representations learned by \acronym{} encode meaningful behavioral structure that goes beyond the predefined activity taxonomy.  Discovered clusters may correspond to known activities of daily living, to finer-grained sub-activities or execution styles, to recurring behavioral routines, or to patterns that do not map cleanly onto any predefined label, thereby potentially surfacing novel insights about resident behavior. This open-ended nature distinguishes clustering from the previous tasks and positions it as an exploratory tool for smart-home sensor event understanding.}

\subsubsection{\added{Task-Specific} Baseline}


\added{ Following the autoencoder-based representation learning approach described in Section \ref{subsubsec:clustering_related} and adopted in prior smart-home research, we use an autoencoder as the baseline for the unsupervised clustering task. Given its established role in the smart-home literature, we adapted the DeepCASAS architecture \cite{liciotti2020sequential} to instantiate the autoencoder. Specifically, we remove the final classification layer of the original DeepCASAS model and attach a symmetric decoder that mirrors the encoder architecture in reverse order. The resulting model is trained end-to-end to reconstruct the input window of sensor events from a compressed latent representation. After training, the decoder is discarded, and the encoder is used to extract embeddings for each window in $\mathcal{W}$. These embeddings are subsequently provided as input to the clustering algorithm. Apart from the modifications required to enable autoencoding, the encoder architecture and hyperparameters are identical to those reported in~\cite{liciotti2020sequential}.}

\subsubsection{\added{Unsupervised Clustering Head}}
\added{Unlike the other downstream tasks, unsupervised clustering does not require a task-specific head. The input to the clustering algorithm is directly the representation $\mathbf{z}_W \in \mathbb{R}^d$ extracted by the encoder for each window $W \in \mathcal{W}$, without any additional trainable component on top.
As a consequence, fine-tuning for this task consists of updating the feature extractor $g_\theta$ itself, using the same dual contrastive loss employed during pre-training, consistent with the other downstream tasks where fine-tuning jointly optimizes the backbone. This choice, shared with \cite{fortes2025coa} among others, reflects the goal of unsupervised clustering, which is to obtain a refined general-purpose representation rather than to optimize for a task-specific objective.} 

\subsubsection{\added{Clustering Algorithm}}
\added{For both \acronym{} and the baseline, we employ HDBSCAN \cite{campello2013density} as the clustering algorithm. HDBSCAN is a density-based hierarchical clustering method that does not require the number of clusters $K$ to be specified a priori, making it particularly well-suited for exploratory analysis where the number of underlying behavioral patterns is unknown. Since the same algorithm is applied to both the representations extracted by \acronym{} and those produced by the baseline encoder, any difference in clustering quality can be attributed solely to the quality of the learned representations rather than to the choice of clustering algorithm. Indeed, the clustering algorithm itself is not the focus of this work, what we aim to evaluate is whether \acronym{} learns representations that capture meaningful behavioral structure in smart-home sensor data.}

\subsubsection{\added{Results}}
\added{Since unsupervised clustering operates without any labels, standard classification metrics are not directly applicable. For evaluation purposes, we leverage the available activity annotations exclusively as an external validation signal to assess the quality of the discovered clusters. Specifically, we adopt cluster purity as our primary evaluation metric, which measures the extent to which each cluster is dominated by windows belonging to a single activity class. In addition to purity, we report the noise percentage (the fraction of windows left unassigned by HDBSCAN) and the number of clusters discovered, both included for completeness. The detailed results are presented in Table \ref{tab:clustering_results} across all datasets.
\acronym{} significantly outperforms the baseline in terms of cluster purity across all datasets, demonstrating that the representations learned by our model capture behavioral structure that aligns more closely with meaningful activity boundaries. Regarding noise percentage and number of discovered clusters, no consistent pattern emerges between the three approaches, confirming that the key differentiating factor is the quality of the learned representations rather than any systematic difference in how HDBSCAN partitions the embedding space. These results further support the effectiveness of our foundation model approach in learning generalizable and semantically meaningful representations of smart-home sensor data, even in a fully unsupervised setting.}

\begin{table}[h]
\centering
\small
\caption{\added{Purity, Noise, and number of discovered patterns (\#)  and training data percentages for each dataset and clustering backbone.}}
\begin{tabular}{ll|ccc|ccc|ccc|ccc}
\toprule
\multirow{2}{*}{\added{\rotatebox{90}{\kern4pt Dataset}}} &

&
\multicolumn{3}{c}{\added{5\%}}
&
\multicolumn{3}{c}{\added{10\%}}
&
\multicolumn{3}{c}{\added{15\%}}
&
\multicolumn{3}{c}{\added{30\%}} \\
\cmidrule(lr){3-5}
\cmidrule(lr){6-8}
\cmidrule(lr){9-11}
\cmidrule(lr){12-14}
& \added{Backbone}
&
\added{Purity} & \added{Noise} & \added{\#}
&
\added{Purity} & \added{Noise} & \added{\#}
&
\added{Purity} & \added{Noise} & \added{\#}
&
\added{Purity} & \added{Noise} & \added{\#}
\\
\midrule

\multirow{3}{*}{\rotatebox{90}{\added{Milan}}}
& \added{DeepCASAS}
& \added{0.24} & \added{0.20} & \added{50.20}
& \added{0.27} & \added{0.21} & \added{53.20}
& \added{0.25} & \added{0.20} & \added{56.40}
& \added{0.29} & \added{0.23} & \added{54.20}
\\
& \added{Chronos-2}
& \added{0.39} & \added{0.29} & \added{68.40}
& \added{0.41} & \added{0.29} & \added{68.00}
& \added{0.45} & \added{0.31} & \added{70.20}
& \added{0.51} & \added{0.30} & \added{71.40}
\\
& \added{DomusFM}
& \added{\textbf{0.62}} & \added{0.30} & \added{71.20}
& \added{\textbf{0.62}} & \added{0.29} & \added{73.40}
& \added{\textbf{0.65}} & \added{0.31} & \added{71.40}
& \added{\textbf{0.71}} & \added{0.33} & \added{72.80}
\\
\midrule

\multirow{3}{*}{\rotatebox{90}{\added{Aruba}}}
& \added{DeepCASAS}
& \added{0.26} & \added{0.29} & \added{70.20}
& \added{0.31} & \added{0.28} & \added{72.60}
& \added{0.38} & \added{0.31} & \added{66.80}
& \added{0.45} & \added{0.28} & \added{64.60}
\\
& \added{Chronos-2}
& \added{0.32} & \added{0.29} & \added{67.20}
& \added{0.39} & \added{0.31} & \added{71.40}
& \added{0.44} & \added{0.32} & \added{74.20}
& \added{0.51} & \added{0.30} & \added{71.40}
\\
& \added{DomusFM}
& \added{\textbf{0.48}} & \added{0.29} & \added{73.40}
& \added{\textbf{0.54}} & \added{0.31} & \added{71.20}
& \added{\textbf{0.57}} & \added{0.32} & \added{74.40}
& \added{\textbf{0.61}} & \added{0.33} & \added{74.20}
\\
\midrule

\multirow{3}{*}{\rotatebox{90}{\added{Kast A}}}
& \added{DeepCASAS}
& \added{0.37} & \added{0.32} & \added{22.20}
& \added{0.39} & \added{0.32} & \added{23.20}
& \added{0.46} & \added{0.31} & \added{25.80}
& \added{0.51} & \added{0.32} & \added{23.80}
\\
& \added{Chronos-2}
& \added{0.46} & \added{0.31} & \added{22.40}
& \added{0.47} & \added{0.31} & \added{22.60}
& \added{0.49} & \added{0.32} & \added{23.20}
& \added{0.54} & \added{0.33} & \added{24.20}
\\
& \added{DomusFM}
& \added{\textbf{0.50}} & \added{0.33} & \added{23.60}
& \added{\textbf{0.51}} & \added{0.32} & \added{25.40}
& \added{\textbf{0.53}} & \added{0.33} & \added{21.80}
& \added{\textbf{0.57}} & \added{0.31} & \added{27.40}
\\
\midrule

\multirow{3}{*}{\rotatebox{90}{\added{Kast C}}}
& \added{DeepCASAS}
& \added{0.37} & \added{0.31} & \added{67.80}
& \added{0.41} & \added{0.29} & \added{72.80}
& \added{0.45} & \added{0.27} & \added{69.80}
& \added{0.51} & \added{0.24} & \added{69.80}
\\
& \added{Chronos-2}
& \added{0.42} & \added{0.27} & \added{61.20}
& \added{0.48} & \added{0.28} & \added{63.00}
& \added{0.51} & \added{0.25} & \added{64.40}
& \added{0.59} & \added{0.25} & \added{62.20}
\\
& \added{DomusFM}
& \added{\textbf{0.63}} & \added{0.24} & \added{57.80}
& \added{\textbf{0.65}} & \added{0.25} & \added{67.80}
& \added{\textbf{0.65}} & \added{0.26} & \added{63.20}
& \added{\textbf{0.72}} & \added{0.22} & \added{69.20}
\\
\midrule

\multirow{3}{*}{\rotatebox{90}{\added{O4H}}}
& \added{DeepCASAS}
& \added{0.49} & \added{0.27} & \added{71.40}
& \added{0.51} & \added{0.31} & \added{73.00}
& \added{0.56} & \added{0.29} & \added{72.40}
& \added{0.58} & \added{0.33} & \added{73.80}
\\
& \added{Chronos-2}
& \added{0.52} & \added{0.30} & \added{65.20}
& \added{0.55} & \added{0.32} & \added{63.40}
& \added{0.62} & \added{0.31} & \added{62.40}
& \added{0.69} & \added{0.32} & \added{61.60}
\\
& \added{DomusFM}
& \added{\textbf{0.74}} & \added{0.25} & \added{64.00}
& \added{\textbf{0.77}} & \added{0.27} & \added{62.00}
& \added{\textbf{0.81}} & \added{0.25} & \added{63.60}
& \added{\textbf{0.84}} & \added{0.23} & \added{63.80}
\\
\midrule

\multirow{3}{*}{\rotatebox{90}{\added{MuRal}}}
& \added{DeepCASAS}
& \added{0.42} & \added{0.30} & \added{70.20}
& \added{0.46} & \added{0.31} & \added{72.40}
& \added{0.49} & \added{0.33} & \added{71.60}
& \added{0.54} & \added{0.33} & \added{75.00}
\\
& \added{Chronos-2}
& \added{0.49} & \added{0.31} & \added{72.20}
& \added{0.51} & \added{0.32} & \added{73.00}
& \added{0.53} & \added{0.31} & \added{74.00}
& \added{0.63} & \added{0.33} & \added{73.20}
\\
& \added{DomusFM}
& \added{\textbf{0.66}} & \added{0.33} & \added{75.20}
& \added{\textbf{0.67}} & \added{0.33} & \added{74.00}
& \added{\textbf{0.71}} & \added{0.31} & \added{75.80}
& \added{\textbf{0.73}} & \added{0.32} & \added{76.20}
\\
\midrule

\multirow{3}{*}{\rotatebox{90}{\added{UCI B}}}
& \added{DeepCASAS}
& \added{0.46} & \added{0.25} & \added{45.00}
& \added{0.47} & \added{0.27} & \added{42.80}
& \added{0.51} & \added{0.23} & \added{43.80}
& \added{0.54} & \added{0.25} & \added{42.60}
\\
& \added{Chronos-2}
& \added{0.49} & \added{0.31} & \added{51.20}
& \added{0.46} & \added{0.33} & \added{48.40}
& \added{0.53} & \added{0.29} & \added{49.20}
& \added{0.57} & \added{0.29} & \added{50.80}
\\
& \added{DomusFM}
& \added{\textbf{0.56}} & \added{0.29} & \added{42.60}
& \added{\textbf{0.56}} & \added{0.27} & \added{45.20}
& \added{\textbf{0.59}} & \added{0.27} & \added{44.40}
& \added{\textbf{0.62}} & \added{0.29} & \added{41.20}
\\
\bottomrule
\end{tabular}
\label{tab:clustering_results}
\end{table}

\subsection{Ablation study} 
\added{To further validate the proposed architecture, we conducted an ablation study focusing on two key components of the foundation model. First, we compared the full \acronym{} architecture with a variant that does not include the contextualized event-level feature extractor, where the task-specific heads operate directly on event-level embeddings rather than on attention-based contextual representations. This ablation aims to quantify the contribution of contextualized event representations and assess whether modeling interactions among events provides additional predictive power beyond independent event-level features. Second, we compared \acronym{} with a version trained without the pre-training stage. This comparison is intended to evaluate the effectiveness of the learned foundation model initialization and determine the extent to which pre-training improves downstream task performance.}

\subsubsection{\added{Remove Contextualization}}
\added{We evaluated the performance of both the full and ablated model (w/o Context) across our three primary downstream tasks: ADL recognition, next-$30$ events prediction, and clustering, considering all the Train Data percentages.
The results, summarized in Tables \ref{tab:ablation_classification}, \ref{tab:ablation_next_30}, and \ref{tab:ablation_clustering}, indicate a consistent decrease in performance across all tasks when the contextualization module is removed. Specifically, for ADL recognition, the drop in F1-score suggests that raw sensor events lack the temporal depth necessary to distinguish between some activities. For next-$30$ events prediction, the reduction in accuracy highlights that encoding the sequential context of the event stream is critical for anticipating future transitions. Similarly, for the unsupervised clustering the lower clustering quality indicates that contextualization is essential for learning discriminative latent representations. Without temporal context, events that occur in similar sequential patterns become less distinguishable, resulting in embeddings with reduced semantic structure and poorer cluster separation.. However, it can be noted that even without the contextualization layer, the results remain competitive and stay above the established baselines. This suggests that, while the contextualization clearly provides a performance boost, the event-level feature extractor alone can still be effective at extracting some meaningful semantic information from individual sensor events.} 

\begin{table}[h]
\centering
\small
\caption{ADL recognition performance (F1-score) for the ablation study, comparing \acronym{} without the contextualization module (w/o Context) against the full architecture (Full) across all seven datasets and different Train Data percentages (5\%, 10\%, 15\%, 30\%).}
\begin{tabular}{lcccccccc}
\toprule
 Train Data \% & \multicolumn{2}{c}{5\%} & \multicolumn{2}{c}{10\%} & \multicolumn{2}{c}{15\%} & \multicolumn{2}{c}{30\%} \\
\midrule
 & w/o Context & Full & w/o Context & Full & w/o Context & Full & w/o Context & Full \\
\midrule
CASAS Milan & 0.67 & \textbf{+0.05} & 0.76 & \textbf{+0.06} & 0.82 & \textbf{+0.04} & 0.90 & \textbf{+0.02} \\
\added{CASAS Aruba} & \added{0.67} & \added{\textbf{+0.09}} & \added{0.71} & \added{\textbf{+0.07}} & \added{0.77} & \added{\textbf{+0.05}} & \added{0.84} & \added{\textbf{+0.07}} \\
\added{Kasteren A} & \added{0.43} & \added{\textbf{+0.05}} & \added{0.51} & \added{\textbf{+0.05}} & \added{0.55} & \added{\textbf{+0.06}} & \added{0.62} & \added{\textbf{+0.06}} \\
Kasteren C & 0.50 & \textbf{+0.09} & 0.59 & \textbf{+0.08} & 0.63 & \textbf{+0.09} & 0.74 & \textbf{+0.07} \\
\added{Orange4Home} & \added{0.81} & \added{\textbf{+0.08}} & \added{0.86} & \added{\textbf{+0.05}} & \added{0.88} & \added{\textbf{+0.06}} & \added{0.91} & \added{\textbf{+0.05}} \\
\added{MuRal} & \added{0.49} & \added{\textbf{+0.11}} & \added{0.59} & \added{\textbf{+0.11}} & \added{0.63} & \added{\textbf{+0.08}} & \added{0.73} & \added{\textbf{+0.07}} \\
\added{UCI B} & \added{0.32} & \added{\textbf{+0.06}} & \added{0.39} & \added{\textbf{+0.07}} & \added{0.40} & \added{\textbf{+0.11}} & \added{0.45} & \added{\textbf{+0.15}} \\
\bottomrule
\end{tabular}
\label{tab:ablation_classification}
\end{table}

\begin{table}[h]
\centering
\small
\caption{Next-$30$ events prediction performance (F1-score) for the ablation study, comparing \acronym{} without the contextualization module (w/o Context) against the full architecture (Full) across all seven datasets and different Train Data percentages (5\%, 10\%, 15\%, 30\%).}
\begin{tabular}{lcccccccc}
\toprule
 Train Data \% & \multicolumn{2}{c}{5\%} & \multicolumn{2}{c}{10\%} & \multicolumn{2}{c}{15\%} & \multicolumn{2}{c}{30\%} \\
\midrule
 & w/o Context & Full & w/o Context & Full & w/o Context & Full & w/o Context & Full \\
\midrule
CASAS Milan & 0.50 & \textbf{+0.07} & 0.53 & \textbf{+0.08} & 0.55 & \textbf{+0.10} & 0.61 & \textbf{+0.12} \\
\added{CASAS Aruba} & \added{0.49} & \added{\textbf{+0.10}} & \added{0.51} & \added{\textbf{+0.11}} & \added{0.56} & \added{\textbf{+0.09}} & \added{0.59} & \added{\textbf{+0.11}} \\
\added{Kasteren A} & \added{0.59} & \added{\textbf{+0.07}} & \added{0.64} & \added{\textbf{+0.10}} & \added{0.69} & \added{\textbf{+0.11}} & \added{0.71} & \added{\textbf{+0.16}} \\
Kasteren C & 0.52 & \textbf{+0.07} & 0.60 & \textbf{+0.09} & 0.65 & \textbf{+0.10} & 0.61 & \textbf{+0.25} \\
\added{Orange4Home} & \added{0.80} & \added{\textbf{+0.06}} & \added{0.83} & \added{\textbf{+0.04}} & \added{0.83} & \added{\textbf{+0.06}} & \added{0.85} & \added{\textbf{+0.04}} \\
\added{MuRal} & \added{0.64} & \added{\textbf{+0.05}} & \added{0.69} & \added{\textbf{+0.06}} & \added{0.73} & \added{\textbf{+0.06}} & \added{0.78} & \added{\textbf{+0.06}} \\
\added{UCI B} & \added{0.70} & \added{\textbf{+0.06}} & \added{0.74} & \added{\textbf{+0.08}} & \added{0.79} & \added{\textbf{+0.07}} & \added{0.81} & \added{\textbf{+0.09}} \\
\bottomrule
\end{tabular}
\label{tab:ablation_next_30}
\end{table}

\begin{table}[h]
\centering
\small
\caption{\added{Unsupervised clustering performance (Purity) for the context ablation study, comparing \acronym{} without the contextualization module (w/o Context) against the full architecture (Full) across all seven datasets and different Train Data percentages (5\%, 10\%, 15\%, 30\%). The Full columns report the gain over the W/O Context version.}}
\begin{tabular}{lcccccccc}
\toprule
 \added{Train Data \%} & \multicolumn{2}{c}{\added{5\%}} & \multicolumn{2}{c}{\added{10\%}} & \multicolumn{2}{c}{\added{15\%}} & \multicolumn{2}{c}{\added{30\%}} \\
\midrule
 & \added{w/o Context} & \added{Full} & \added{w/o Context} & \added{Full} & \added{w/o Context} & \added{Full} & \added{w/o Context} & \added{Full} \\
\midrule
\added{CASAS Milan}   & \added{0.51} & \added{\textbf{+0.11}} & \added{0.52} & \added{\textbf{+0.10}} & \added{0.56} & \added{\textbf{+0.09}} & \added{0.59} & \added{\textbf{+0.12}} \\
\added{CASAS Aruba}   & \added{0.40} & \added{\textbf{+0.08}} & \added{0.41} & \added{\textbf{+0.13}} & \added{0.45} & \added{\textbf{+0.12}} & \added{0.49} & \added{\textbf{+0.12}} \\
\added{Kasteren A}  & \added{0.39} & \added{\textbf{+0.11}} & \added{0.43} & \added{\textbf{+0.08}} & \added{0.46} & \added{\textbf{+0.07}} & \added{0.49} & \added{\textbf{+0.08}} \\
\added{Kasteren C}  & \added{0.52} & \added{\textbf{+0.11}} & \added{0.53} & \added{\textbf{+0.12}} & \added{0.56} & \added{\textbf{+0.09}} & \added{0.61} & \added{\textbf{+0.11}} \\
\added{Orange4Home}     & \added{0.69} & \added{\textbf{+0.05}} & \added{0.71} & \added{\textbf{+0.06}} & \added{0.75} & \added{\textbf{+0.06}} & \added{0.79} & \added{\textbf{+0.05}} \\
\added{MuRal}   & \added{0.55} & \added{\textbf{+0.11}} & \added{0.58} & \added{\textbf{+0.09}} & \added{0.63} & \added{\textbf{+0.08}} & \added{0.67} & \added{\textbf{+0.06}} \\
\added{UCI B}   & \added{0.44} & \added{\textbf{+0.12}} & \added{0.47} & \added{\textbf{+0.09}} & \added{0.51} & \added{\textbf{+0.08}} & \added{0.55} & \added{\textbf{+0.07}} \\
\bottomrule
\end{tabular}
\label{tab:ablation_clustering}
\end{table}

\subsubsection{\added{Remove Pre-training}}
\added{We evaluated the performance of both the full and ablated model (w/o Pretrain) across our three primary downstream tasks: ADL recognition, next-30 events prediction, and clustering, considering all the Train Data percentages.
The results, summarized in Tables \ref{tab:no_pt_adl}, \ref{tab:no_pt_n30}, and \ref{tab:no_pt_clustering}, indicate a consistent decrease in performance across all tasks without the pre-training.
For the ADL recognition task, the performance gain introduced by pre-training is comparatively less pronounced. We hypothesize that this is due to the nature of the task, which largely reduces to a conventional supervised classification problem when some labeled data is available for fine-tuning. In this setting, even without pre-training, the Transformer-based architecture, combined with the semantic representations extracted from the LLM, is already capable of achieving strong performance with relatively limited training data. As a result, while pre-training still provides a consistent improvement, its impact is less substantial than in the other evaluated tasks.
In contrast, for both the next-30 events prediction and the unsupervised clustering task, the benefit of pre-training is significantly more evident. In these scenarios, the w/o Pretrain variant exhibits a considerably larger performance degradation, highlighting the critical role of pre-training in learning transferable representations. This effect is particularly pronounced in settings that require either long-horizon temporal reasoning or structure discovery without explicit supervision. Overall, these results further support the effectiveness of our pre-trained foundation model in learning general-purpose representations that transfer across diverse downstream tasks.}

\begin{table}[h]
\centering
\small
\caption{\added{ADL recognition performance (F1-score) for the ablation study, comparing \acronym{} without the pretrain (w/o Pretrain) against the full pipeline (Full) across all seven datasets and different Train Data percentages (5\%, 10\%, 15\%, 30\%).}}
\begin{tabular}{lcccccccc}
\toprule
 \added{Train Data \%} & \multicolumn{2}{c}{\added{5\%}} & \multicolumn{2}{c}{\added{10\%}} & \multicolumn{2}{c}{\added{15\%}} & \multicolumn{2}{c}{\added{30\%}} \\
\midrule
 & \added{w/o Pretrain} & \added{Full} & \added{w/o Pretrain} & \added{Full} & \added{w/o Pretrain} & \added{Full} & \added{w/o Pretrain} & \added{Full} \\
\midrule
\added{CASAS Milan} & \added{0.58} & \added{\textbf{+0.14}} & \added{0.71} & \added{\textbf{+0.11}} & \added{0.82} & \added{\textbf{+0.04}} & \added{0.87} & \added{\textbf{+0.05}} \\
\added{CASAS Aruba} & \added{0.65} & \added{\textbf{+0.11}} & \added{0.69} & \added{\textbf{+0.09}} & \added{0.74} & \added{\textbf{+0.08}} & \added{0.80} & \added{\textbf{+0.11}} \\
\added{Kasteren A} & \added{0.44} & \added{\textbf{+0.04}} & \added{0.47} & \added{\textbf{+0.09}} & \added{0.49} & \added{\textbf{+0.12}} & \added{0.57} & \added{\textbf{+0.11}} \\
\added{Kasteren C} & \added{0.44} & \added{\textbf{+0.15}} & \added{0.52} & \added{\textbf{+0.15}} & \added{0.60} & \added{\textbf{+0.12}} & \added{0.75} & \added{\textbf{+0.06}} \\
\added{Orange4Home} & \added{0.84} & \added{\textbf{+0.05}} & \added{0.85} & \added{\textbf{+0.06}} & \added{0.88} & \added{\textbf{+0.06}} & \added{0.92} & \added{\textbf{+0.04}} \\
\added{MuRal} & \added{0.41} & \added{\textbf{+0.19}} & \added{0.58} & \added{\textbf{+0.12}} & \added{0.65} & \added{\textbf{+0.06}} & \added{0.76} & \added{\textbf{+0.04}} \\
\added{UCI B} & \added{0.30} & \added{\textbf{+0.08}} & \added{0.32} & \added{\textbf{+0.14}} & \added{0.33} & \added{\textbf{+0.18}} & \added{0.36} & \added{\textbf{+0.24}} \\
\bottomrule
\end{tabular}
\label{tab:no_pt_adl}
\end{table}

\begin{table}[h]
\centering
\small
\caption{\added{Next-$30$ events prediction performance (F1-score) for the ablation study, comparing \acronym{} without the pretrain (w/o Pretrain) against the full pipeline (Full) across all seven datasets and different Train Data percentages (5\%, 10\%, 15\%, 30\%).}}
\begin{tabular}{lcccccccc}
\toprule
 \added{Train Data \%} & \multicolumn{2}{c}{\added{5\%}} & \multicolumn{2}{c}{\added{10\%}} & \multicolumn{2}{c}{\added{15\%}} & \multicolumn{2}{c}{\added{30\%}} \\
\midrule
 & \added{w/o Pretrain} & \added{Full} & \added{w/o Pretrain} & \added{Full} & \added{w/o Pretrain} & \added{Full} & \added{w/o Pretrain} & \added{Full} \\
\midrule
\added{CASAS Milan} & \added{0.46} & \added{\textbf{+0.11}} & \added{0.48} & \added{\textbf{+0.13}} & \added{0.50} & \added{\textbf{+0.15}} & \added{0.55} & \added{\textbf{+0.18}} \\
\added{CASAS Aruba} & \added{0.49} & \added{\textbf{+0.10}} & \added{0.51} & \added{\textbf{+0.11}} & \added{0.53} & \added{\textbf{+0.12}} & \added{0.59} & \added{\textbf{+0.11}} \\
\added{Kasteren A} & \added{0.51} & \added{\textbf{+0.15}} & \added{0.56} & \added{\textbf{+0.18}} & \added{0.58} & \added{\textbf{+0.22}} & \added{0.67} & \added{\textbf{+0.20}} \\
\added{Kasteren C} & \added{0.41} & \added{\textbf{+0.18}} & \added{0.47} & \added{\textbf{+0.22}} & \added{0.54} & \added{\textbf{+0.21}} & \added{0.66} & \added{\textbf{+0.20}} \\
\added{Orange4Home} & \added{0.65} & \added{\textbf{+0.21}} & \added{0.66} & \added{\textbf{+0.21}} & \added{0.70} & \added{\textbf{+0.19}} & \added{0.74} & \added{\textbf{+0.15}} \\
\added{MuRal} & \added{0.50} & \added{\textbf{+0.19}} & \added{0.53} & \added{\textbf{+0.22}} & \added{0.54} & \added{\textbf{+0.25}} & \added{0.60} & \added{\textbf{+0.24}} \\
\added{UCI B} & \added{0.53} & \added{\textbf{+0.23}} & \added{0.56} & \added{\textbf{+0.26}} & \added{0.58} & \added{\textbf{+0.28}} & \added{0.65} & \added{\textbf{+0.25}} \\
\bottomrule
\end{tabular}
\label{tab:no_pt_n30}
\end{table}

\begin{table}[h]
\centering
\small
\caption{\added{Unsupervised clustering performance (Purity) for the ablation study, comparing \acronym{} without pretraining (w/o Pretrain) against the full architecture (Full) across all seven datasets and different Train Data percentages (5\%, 10\%, 15\%, 30\%). The Full columns report the gain over the w/o Pretrain version.}}
\begin{tabular}{lcccccccc}
\toprule
 \added{Train Data \%} & \multicolumn{2}{c}{\added{5\%}} & \multicolumn{2}{c}{\added{10\%}} & \multicolumn{2}{c}{\added{15\%}} & \multicolumn{2}{c}{\added{30\%}} \\
\midrule
 & \added{w/o Pretrain} & \added{Full} & \added{w/o Pretrain} & \added{Full} & \added{w/o Pretrain} & \added{Full} & \added{w/o Pretrain} & \added{Full} \\
\midrule
\added{CASAS Milan}   & \added{0.35} & \added{\textbf{+0.27}} & \added{0.36} & \added{\textbf{+0.26}} & \added{0.38} & \added{\textbf{+0.27}} & \added{0.45} & \added{\textbf{+0.26}} \\
\added{CASAS Aruba}   & \added{0.30} & \added{\textbf{+0.18}} & \added{0.32} & \added{\textbf{+0.22}} & \added{0.40} & \added{\textbf{+0.17}} & \added{0.41} & \added{\textbf{+0.20}} \\
\added{Kasteren A}    & \added{0.44} & \added{\textbf{+0.06}} & \added{0.43} & \added{\textbf{+0.08}} & \added{0.45} & \added{\textbf{+0.08}} & \added{0.49} & \added{\textbf{+0.08}} \\
\added{Kasteren C}    & \added{0.50} & \added{\textbf{+0.13}} & \added{0.53} & \added{\textbf{+0.12}} & \added{0.54} & \added{\textbf{+0.11}} & \added{0.59} & \added{\textbf{+0.13}} \\
\added{Orange4Home}   & \added{0.52} & \added{\textbf{+0.22}} & \added{0.54} & \added{\textbf{+0.23}} & \added{0.62} & \added{\textbf{+0.19}} & \added{0.66} & \added{\textbf{+0.18}} \\
\added{MuRal}         & \added{0.46} & \added{\textbf{+0.20}} & \added{0.49} & \added{\textbf{+0.18}} & \added{0.56} & \added{\textbf{+0.15}} & \added{0.58} & \added{\textbf{+0.15}} \\
\added{UCI B}         & \added{0.36} & \added{\textbf{+0.20}} & \added{0.41} & \added{\textbf{+0.15}} & \added{0.41} & \added{\textbf{+0.18}} & \added{0.48} & \added{\textbf{+0.14}} \\
\bottomrule
\end{tabular}
\label{tab:no_pt_clustering}
\end{table}

\subsection{\added{Resource Requirements and Model Features for Edge Deployment}}
\label{subsec:real_life_deployment}
\added{A key design objective of \acronym{} is to ensure practicality and feasibility in real-world smart-home deployments. Unlike many recent approaches that rely on large-scale models or external cloud-based services, \acronym{} is explicitly engineered to be lightweight and computationally efficient, making it suitable for execution on edge devices commonly available in smart-home infrastructures.}


\added{We conducted an experiment installing the complete pretrained \acronym{} on a MiniPC equipped with an Intel Celeron J4125 quad-core CPU, 8 GB of DDR RAM, and a 256 GB SSD, i.e., an inexpensive device that could be considered representative of standard smart-home gateways.
Our current pretrained model has a total memory footprint of less than 500 MB, hence compatible with low-end edge devices, like the one used in our experiment. 
We limited our experiment to the downstream tasks of ADL recognition and next-\textit{k} events prediction, and to the datasets CASAS Milan and Kasteren C, representative of different home layouts and sensor event densities.}

\added{The edge device was sufficiently powerful to perform the fine-tuning with 5\% of the training data in a few hours.
The average inference time required to process a single window of sensor events was $9.9\pm1$ milliseconds for ADL recognition and $9.6\pm1$ milliseconds for next-\textit{k} events prediction.}
\added{Based on these results, \acronym{} is expected to operate efficiently on standard smart-home gateways.
}


\added{This lightweight design is largely enabled by the event-based nature of \acronym{}. By representing sensor activity as discrete events rather than continuous streams, the resulting data is considerably sparser, allowing the model to achieve comparable performance with fewer parameters and, consequently, a smaller memory footprint. This makes the event-based approach not only effective for modeling sensor data but also inherently promising from a deployability standpoint.}

Edge deployment offers several important advantages over cloud-based solutions. First, it eliminates the need to transmit sensitive in-home activity data to external servers or third-party APIs, thereby significantly reducing privacy risks and aligning with the stringent privacy requirements of domestic environments. Second, it avoids recurring costs and latency associated with commercial API-based services, making the solution economically sustainable for long-term deployment. Finally, local inference improves system robustness and scalability, as performance is not affected by network connectivity issues or external service availability.

Beyond computational efficiency, the experimental results presented in 
this paper
demonstrate that \acronym{} can be effectively fine-tuned with limited labeled data while maintaining strong performance across previously unseen environments. In particular, the leave-one-dataset-out evaluation shows that the model generalizes well to new homes, different sensor configurations, and unseen occupants, despite substantial variability in layouts, sensing infrastructures, and activity distributions. \added{This strong transfer capability is enabled by the semantic nature of the embeddings extracted by the LLM, which capture functional relationships beyond dataset-specific sensor representations, making the model more robust to differences in sensor types, naming conventions, and even missing attributes.} 

Overall, these properties ensure that \acronym{} can be realistically deployed in real-world smart-home scenarios at low cost, with strong privacy guarantees and without reliance on external services. 

\section{Discussion}

\subsection{Comparison with Zero-shot LLM Approaches}
As discussed in Section \ref{subsec:fm_4_har}, several recent works have proposed the use of large LLMs to address the ADL recognition problem in smart-home environments. These approaches typically frame activity recognition as a language understanding task by transforming sensor event streams into textual representations and leveraging the reasoning capabilities of pretrained LLMs. Importantly, these methods are limited to ADL recognition and do not extend to other smart-home tasks such as next-\textit{k} event prediction.

A key advantage of LLM-based approaches is their ability to operate in a fully zero-shot setting, requiring no labeled training data. While this property is attractive, it is worth noting that, as discussed earlier in the paper, acquiring a limited amount of labeled data in smart-home scenarios is often realistic (e.g., during system installation or short calibration phases). As such, evaluating the trade-off between zero-shot performance and lightweight fine-tuning is crucial for assessing real-world applicability.

To this end, we compared \acronym{} against a representative zero-shot LLM-based approach proposed in \cite{fiori2026improving}, which operates in a setting closely aligned with ours. Specifically, it performs ADL recognition using event-based segmentation with a fixed window length of 30 events, matching our experimental configuration. In addition to reproducing the original ADL recognition setup, we extended this approach to evaluate its performance on the next-\textit{30} events prediction task, enabling a broader comparison across tasks.

The quantitative comparison is reported in Tables \ref{tab:llm_comparison_clf} (ADL recognition) and \ref{tab:llm_comparison_next_k} (next-\textit{30} events prediction). The results show that \acronym{} consistently outperforms the zero-shot LLM-based approach on both datasets considered, even when fine-tuned using the smallest percentage of labeled data. This highlights the effectiveness of our foundation model in leveraging minimal supervision to achieve superior performance compared to purely zero-shot solutions.

It is important to emphasize that the LLM-based approach was evaluated strictly in its zero-shot configuration. A more exhaustive comparison could investigate few-shot prompting strategies, where labeled training instances are provided as in-context examples to the LLM. While such an analysis is beyond the scope of this work, it would come at the cost of increased prompt complexity, memory usage, and inference latency.

For this comparison, we used the best-performing model reported in \cite{fiori2026improving}, namely Gemma 27B, deployed locally on a Linux-based machine equipped with an NVIDIA A100 PCIe GPU with 80 GB of dedicated VRAM.

The LLM-based solution exhibited substantial computational overhead. The average inference times for ADL recognition were $8 \pm 3$ seconds. For the next-\textit{k} events prediction task, inference times further increased to $15 \pm 5$ seconds. In addition, the model requires 18 GB of memory. These computational and memory requirements are orders of magnitude greater than those of \acronym{}, shown in Section \ref{subsec:real_life_deployment}, a comparison of computational requirements is show in Table \ref{tab:computational_comparison}.

\begin{table}[h]
\centering
\caption{Computational Requirements Comparison: \acronym{} demonstrates significantly lower memory footprint and faster inference times compared to the LLM-based solution, enabling deployment on standard smart-home gateways without specialized hardware.}
\label{tab:computational_comparison}
\begin{tabular}{lcc}
\toprule
\textbf{Metric} & \textbf{\acronym{}} & \textbf{LLM-based Solution} \\
\midrule
\textit{Memory Usage} & $<$ 500 MB & 18 GB \\
\midrule
\multicolumn{3}{l}{\textit{Inference Time}} \\
ADL recognition & $9.9 \pm 1$ ms & $8 \pm 3$ s \\
Next-\textit{k} Events Prediction & $9.6 \pm 1$ ms & $15 \pm 5$ s \\
\midrule
Deployment Requirements & Standard gateway & Specialized hardware \\
\bottomrule
\end{tabular}
\end{table}

\begin{table}[t]
\centering
\caption{ADL recognition performance (Weighted F1-score) comparing \acronym{} against the zero-shot LLM approach.}
\label{tab:llm_comparison_clf}
\begin{tabular}{|l|c|cccc|}
\hline
& LLM & \multicolumn{4}{c|}{\acronym} \\
& Zero-shot & 5\% & 10\% & 15\%  & 30\% \\
\hline
CASAS Milan & 0.56 & 0.72 & 0.82 & 0.86 & 0.92 \\
\hline
Kasteren C & 0.48 & 0.59 & 0.67 & 0.72 & 0.81 \\
\hline
\end{tabular}
\end{table}

\begin{table}[t]
\centering
\caption{Next-$30$ events prediction performance (F1-score) comparing \acronym{} against the zero-shot LLM approach.}
\label{tab:llm_comparison_next_k}
\begin{tabular}{|l|c|cccc|}
\hline
& LLM & \multicolumn{4}{c|}{\acronym} \\
& Zero-shot & 5\% & 10\% & 15\%  & 30\% \\  
\hline
CASAS Milan & 0.45 & 0.57 & 0.61 & 0.65 & 0.73 \\
\hline
Kasteren C & 0.49 & 0.59 & 0.69 & 0.75 & 0.86 \\
\hline
\end{tabular}
\end{table}

\subsection{\added{Impact of Density on Next Events Prediction}}
\added{An important consideration when interpreting the next-\textit{k} events prediction results is that, across the seven datasets used in this work, the temporal span covered by a fixed number of events can vary substantially. This variability stems from differences in event density, i.e., how temporally concentrated sensor events are within a dataset. Some environments produce frequent, closely spaced events (high density), while others generate sparse events separated by long intervals (low density). While we believe that predicting a fixed number of future events remains a meaningful and practically useful task formulation, we wanted to complement our evaluation by also assessing performance when the prediction is instead defined over a fixed time frame, ensuring a consistent and comparable time window across datasets with different density profiles.}

\added{To this end, we conducted an additional experiment in which, instead of fixing the number of events to predict, we fixed a time span and predicted all events expected to occur within it. We selected a 5-minute window, as it represents a temporal granularity commonly relevant for behavioral analysis.  We evaluated this formulation on two datasets representative of the two density profiles observed in our data: CASAS Milan (high density) and Kasteren C (low density). }

\added{The results, reported in Table \ref{tab:next-5 minutes}, show that while \acronym{}'s performance is slightly lower under this time-based formulation compared to the fixed-event-count formulation, across both datasets and all Train Data percentages, it still consistently outperforms both baselines and maintains satisfying predictive performance. These findings suggest that \acronym{}'s learned representations generalize robustly regardless of whether the prediction is defined in terms of event count or elapsed time, further supporting its applicability to real-world scenarios where behaviorally meaningful time frames, rather than arbitrary event counts, are sometimes of greater practical interest.}

\begin{table}[h]
\centering
\footnotesize
\caption{Next-events prediction performance (F1-score) over a fixed 5-minute time window, comparing DomusFM
against the GPT-2 and Chronos baselines on CASAS Milan and Kasteren C, for Train Data percentages of 5\% and 10\%.
Results are averaged over 5-fold cross-validation.}

\begin{tabular}{lcccccc}
\toprule
\added{Train Data \%} & \multicolumn{3}{c}{\added{5\%}} & \multicolumn{3}{c}{\added{10\%}} \\
\midrule
 & \added{GPT-2} & \added{Chronos} & \added{\acronym} & \added{GPT-2} & \added{Chronos} & \added{\acronym} \\
\midrule
\added{CASAS Milan} & \added{0.36} & \added{0.39} & \added{\textbf{0.55}} & \added{0.43} & \added{0.47} & \added{\textbf{0.58}} \\
\added{Kasteren C} & \added{0.40} & \added{0.41} & \added{\textbf{0.52}} & \added{0.42} & \added{0.46} & \added{\textbf{0.63}} \\
\bottomrule
\end{tabular}

\vspace{1em}

\begin{tabular}{lcccccc}
\toprule
\added{Train Data \%} & \multicolumn{3}{c}{\added{15\%}} & \multicolumn{3}{c}{\added{30\%}} \\
\midrule
 & \added{GPT-2} & \added{Chronos} & \added{\acronym} & \added{GPT-2} & \added{Chronos} & \added{\acronym} \\
\midrule
\added{CASAS Milan} & \added{0.51} & \added{0.57} & \added{\textbf{0.61}} & \added{0.59} & \added{0.63} & \added{\textbf{0.71}} \\
\added{Kasteren C} & \added{0.43} & \added{0.52} & \added{\textbf{0.71}} & \added{0.65} & \added{0.68} & \added{\textbf{0.82}} \\

\bottomrule
\end{tabular}

\label{tab:next-5 minutes}
\end{table}

\subsection{Interpreting the Discovered Clusters}
\label{subsec:cluster_interpretation}

The clustering experiments in Section~\ref{subsec:unsupervised_clustering} assess cluster quality through purity, using activity labels only as an external validation signal. Purity, however, does not reveal what each cluster represents. Since no labels are used to form the clusters, interpreting them requires inspecting their content. A cluster can be characterized by selecting the windows closest to its center in the embedding space (i.e., its most representative members) and mapping them back to the underlying sensor event sequences. 

\begin{figure}
    \centering
    \includegraphics[width=0.9\linewidth]{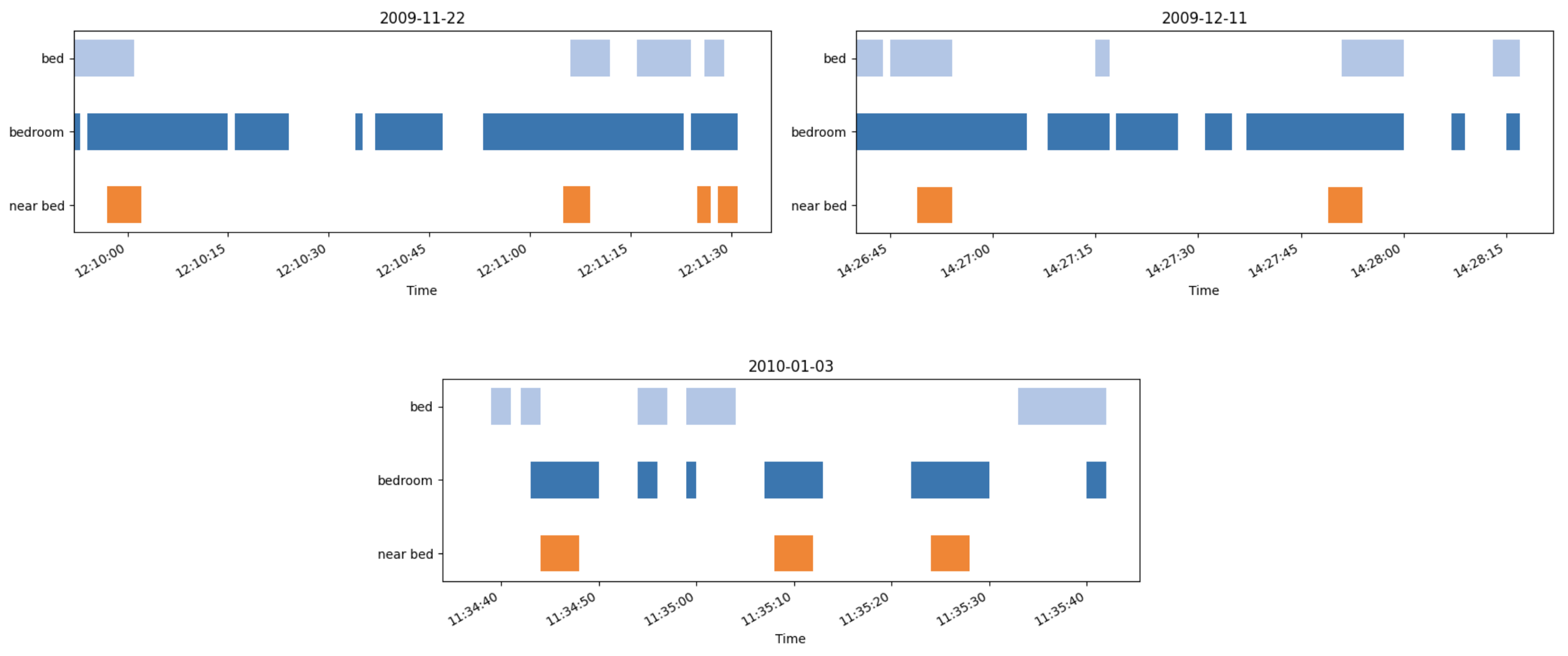}
    \caption{Three windows from the same CASAS Milan cluster. Bars show sensor activations (ON to OFF). Despite coming from different days, all windows show repeated bedroom and near-bed activity around midday.}
    \label{fig:cluster_example}
\end{figure}

As an example, Figure~\ref{fig:cluster_example} shows three windows belonging to the same cluster discovered on CASAS Milan, randomly sampled among the 15 windows closest to the cluster centroid. The windows come from different days, and each spans roughly one to one and a half minutes. Despite this, they share a recognizable pattern: the bedroom sensor remains active almost continuously, interleaved with short, repeated activations of the bed and near-bed sensors, and all three occur around midday. The recurrence of this pattern across different days suggests that the cluster captures a recurring behavior rather than a one-off episode, even though no activity label was used to form it. The group of selected windows is fed altogether to a semantic engine (e.g., an LLM, an ontology, a human expert...). For each event in these windows, the semantic engine receives the full set of semantic attributes modeled by \acronym{}: which sensor fired (house item, sensor type, and room) and when (day of the week and time of day).  Aggregating these descriptions across the representative windows yields a semantic description of the cluster, e.g., ``Moving in the bedroom around noon''.

The granularity of the discovered patterns is not fixed by \acronym{}, but depends on two factors. The first is the sensing infrastructure: a home instrumented mostly with motion sensors supports patterns of movement and room occupancy, whereas a home with object-level sensors also supports patterns of object interaction. The second is the segmentation: Intuitively, the more events in a window, the coarser the granularity of the semantic concepts discoverable by a semantic engine based on its latent-space remapping.  For example, in CASAS Milan, given the specific infrastructure and the windows of $N=30$ events, the patterns last at most a few minutes. Hence, a semantic engine can infer only semantic concepts of relatively fine granularity. Inferring semantic concepts of larger temporal granularity with the same infrastructure, such as full daily routines, would require larger windows or a second-level analysis over groups of clusters.

\subsection{\added{Beyond Transformer-Based Models for Language}}

\added{At first glance, the use of Transformer-based models over smart-home event sequences may invite an analogy with models for language, where token representations are learned from their surrounding context.}
\added{Indeed, \acronym{} shares this general idea of contextual representation learning: individual smart-home events can be regarded as token-like units whose representations are shaped by neighboring events. However, this analogy is inherently limited. Unlike linguistic tokens, which are usually treated as discrete symbolic units, smart-home events are heterogeneous and multidimensional: they jointly encode semantic attributes, such as sensor type and state; spatial attributes, such as room and object location; and temporal attributes, such as the time of occurrence and the elapsed time between interactions. Learning effective event representations, therefore, requires integrating multiple sources of information rather than embedding a single symbolic identifier.}

\added{
A further difference concerns the role of temporal structure. While recent NLP models primarily exploit token order to model syntactic and semantic dependencies, smart-home behavior depends on both event ordering and the temporal gaps between events. Two sequences with the same ordering but different inter-event times may correspond to different behaviors, activity levels, or routines. Conversely, many activities exhibit a partial-order structure: several event permutations may instantiate the same underlying behavior. For instance, during meal preparation, opening a cabinet, using a countertop appliance, and interacting with the fridge may occur in different orders without necessarily changing the activity label. Hence, instead of using standard positional encoding like Transformer-based language models, \acronym{} uses a richer temporal encoding that captures irregular inter-event intervals rather than mere sequential position.} 

\added{
Finally, the data regime differs substantially from that of recent NLP models. They benefit from massive, diverse, and widely available text corpora, whereas smart-home datasets are typically small, fragmented, privacy-sensitive, environment-specific, and expensive to collect and annotate. This scarcity makes it difficult to scale \acronym{} through data volume alone and increases the importance of self-supervised learning, transfer across homes, domain adaptation, and representations that remain robust under limited supervision, sensor heterogeneity, and missing observations.}

\added{This perspective makes explicit the unique nature of smart-home events and clarifies why their representation requires a domain-specific notion of context.}

\subsection{Limitations}
\label{subsec:limitations}

While \acronym{} demonstrates strong performance across diverse smart-home environments, some limitations warrant discussion. 

\subsubsection{Single-Occupancy Scenarios} Our approach focuses on environments with either single occupancy or perfect data association, where each sensor activation can be attributed to a specific individual. While single-occupant scenarios are realistic (e.g., elderly individuals living alone), perfect data association in multi-occupant settings remains an open challenge \cite{arrotta2025multi}. However, we believe \acronym{}'s rich semantic representations could contribute to addressing this problem. Future work will explore leveraging our learned representations for improved data association in multi-occupant environments.

\subsubsection{Binarization Requirement} Our framework requires continuous sensor data to be converted into binary semantic states. While effective in capturing semantically meaningful information, this necessitates preprocessing and may result in information loss. Future work will explore architectures that directly integrate continuous sensor streams alongside binary events, reducing preprocessing requirements while preserving temporal dynamics and semantics.

\subsubsection{Zero-Shot Generalization} \acronym{} does not yet operate in a zero-shot fashion. Some cross-dataset approaches achieve zero-shot performance by mapping all activities to a common set, but this reduces activity complexity and weakens the evaluation of true generalization. In contrast, \acronym{} preserves environment-specific activity definitions and adapts with minimal supervision (5\% labeled data) without activity mapping. Future work will investigate natural language supervision techniques to enable fully zero-shot operation while maintaining the original activity sets.

\subsubsection{\added{Task Coverage}} \added{We evaluated three downstream tasks that we consider particularly 
relevant for behavioral monitoring, but many more would deserve investigation and experimental testing,
%
including, for instance, anomaly detection, behavior change detection, and occupancy prediction.}

\section{Conclusion and Future Work}
This paper introduces \acronym{}, the first foundation model specifically designed and pretrained \added{for event-based behavioral monitoring in smart homes}. We formalized the conceptual framework for event-level and contextualized representations, designed a novel architecture integrating semantic understanding with specialized temporal encoders, and demonstrated through rigorous leave-one-dataset-out evaluation across seven datasets that \acronym{} substantially outperforms state-of-the-art baselines with minimal supervision (5\% labeled data).
\acronym{} represents a paradigm shift from task-specific models to a general-purpose foundation model that rapidly adapts to diverse environments, infrastructures, and users. 
Its lightweight architecture enables edge deployment, addressing critical privacy and latency concerns for practical smart-home systems. 
By demonstrating effective transfer learning despite data scarcity, this work opens new possibilities for the widespread adoption of intelligent home systems.
Future work, as outlined in Section \ref{subsec:limitations}, will investigate how to extend the model to multi-occupant environments, integrate continuous sensor streams directly, enable fully zero-shot generalization, and validate adaptability across broader smart-home tasks. \added{With fewer than 37 million parameters and a memory footprint of under 500 MB, DomusFM is designed to be compact; we will also conduct a scaling analysis to assess whether increasing model capacity can further improve performance while preserving this efficiency.}

\renewcommand{\acksname}{Acknowledgment}
\begin{acks}
This paper is supported by FAIR (Future Artificial Intelligence  Research) project, funded by the NextGenerationEU program.
\end{acks}

\interlinepenalty=10000

\bibliographystyle{ACM-Reference-Format}
\bibliography{sample-base}

\appendix
\section{\added{Background and Data Processing}}
\label{sec:preliminaries_data_processing}

\label{apx:preprocessing}
A smart home is a home environment enriched with sensing, actuation, communication, and computation capabilities that permit it to adapt to users' behaviors \cite{cicirelli2016design}.
We consider smart-home scenarios involving one or more sensorized environments where each sensor activation can be accurately attributed to a specific individual. This attribution may be achieved through either single-occupancy environments or advanced sensing systems capable of identifying the person responsible for each sensor interaction \cite{arrotta2025multi}.

Formally, let $S = \{s_0, s_1, \ldots, s_M\}$ be the set of $M$ sensors deployed in the environment. Each sensor $s_i$ is characterized by its ID, an associated monitored object ($s_i^{HouseItem}$), the installation location ($s_i^{Room}$), and a specific modality ($s_i^{Type}$). For improved interoperability, attribute names can be drawn from existing smart-home ontologies \cite{katz1970progress, daniele2015created, janowicz2019sosa, civitarese2019polaris}. However, our system is designed in an open-world fashion, imposing no strict constraints on prior knowledge engineering.

Sensors are categorized into two kinds based on the type of data they produce:

\begin{itemize}

\item \textbf{Binary sensors}, such as contact switches or motion detectors, directly output discrete states, like OPEN/CLOSE or ON/OFF, that have an immediate and explicit semantic interpretation related to human actions. For an ideal binary sensor, the event sequence must alternate. Formally, for any two consecutive events $e = \langle t, s, \sigma \rangle$ and $e' = \langle t', s', \sigma' \rangle$ with $t < t'$ and no intermediate events from sensor $s$, it must hold that:
$$
\sigma \neq \sigma'
$$
Violations of this condition typically indicate sensor malfunctions, network interruptions, or infrastructure anomalies. Such duplicate events are safely removed during the data cleaning process prior to segmentation.
\item \textbf{Continuous sensors}, such as temperature sensors or power meters, produce a stream of real values ($\mathcal{V} \subseteq \mathbb{R}$). Contrary to HAR systems based on wearable devices, the smart-home literature for behavioral modeling rarely relies on continuous sensor data in its raw form. Nonetheless, various strategies have been proposed, such as summarizing them using statistical features \cite{huang2023human}, or, most commonly, converting continuous values into discrete semantic states \cite{liciotti2020sequential, das2023explainable, arrotta2022dexar}. Despite methodological differences, these approaches share the principle of abstracting continuous data into discrete representations.
\end{itemize}

More formally, while binary sensors natively admit two mutually exclusive states  $\mathcal{V} = \{1,0\}$, continuous sensors are converted into discrete semantic states to capture behaviorally relevant information and enhance interpretability. Let $x : T \to \mathcal{V} \subseteq \mathbb{R}$ be a continuous sensor and $\mathcal{P} = \{I_1, \ldots, I_m\}$ a set of $m$ semantic states (e.g., high temperature intervals). The associated binary states are given by the indicator function:
$$
\sigma_j(t) = \mathbf{1}_{\{x(t) \in I_j\}}, \quad j=1,\dots,m
$$
so that each $\sigma_j : T \to \{0,1\}$ defines a discrete binary state stream \cite{zhang2020fusion,song2021secure,hu2022kalman,han2024local}. This abstraction captures the behaviorally relevant information needed for analysis while improving interpretability, as raw numerical values often lack clear semantic meaning and fine-grained temporal variations may not convey useful behavioral cues.

\begin{figure}
    \centering

    \begin{subfigure}{0.44\linewidth}
        \centering
        \includegraphics[width=\linewidth]{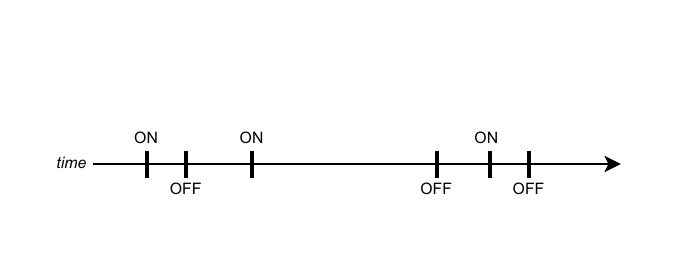}
        \caption{An example of the event stream produced by a binary sensor.}
        \label{fig:binary_sensor}
    \end{subfigure}
    \hfill
    \begin{subfigure}{0.44\linewidth}
        \centering
        \includegraphics[width=\linewidth]{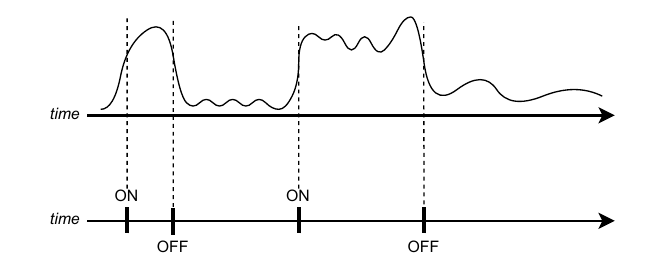}
        \caption{The event stream of a continuous sensor can be abstracted into binary states.}
        \label{fig:continuous_to_binary}
    \end{subfigure}
    \captionsetup{singlelinecheck=false}
    \caption{Event streams from the two different kinds of sensors.}
    \label{fig:sensors}
\end{figure}

Figure \ref{fig:binary_sensor} shows the stream of events of a binary sensor (e.g., a magnetic switch). Figure \ref{fig:continuous_to_binary} instead shows how it is possible to extract high-level binary states from a continuous sensor (e.g., HIGH TEMPERATURE states) from raw (e.g., temperature) data. 

\end{document}